\documentclass[review]{elsarticle}

\usepackage[a4paper, total={6in, 9in}]{geometry}

\usepackage[dvipsnames]{xcolor}

\usepackage{lineno,hyperref}
\modulolinenumbers[5]

\usepackage[utf8]{inputenc}
\usepackage{times}
\usepackage{graphicx}
\usepackage{amssymb}
\usepackage{amsmath}
\usepackage{multirow}
\usepackage{adjustbox}	% - Compressing tables
\usepackage{booktabs}
\usepackage{url}
\usepackage{hyperref}
\usepackage{algorithm}
\usepackage{algorithmic}
\usepackage{tabularx}
\usepackage{float}

\graphicspath{{figs/}}
\DeclareGraphicsExtensions{.pdf,.png,.jpg}

\usepackage{comment}

\usepackage{caption}
\usepackage{subcaption}

\journal{Engineering Applications of Artificial Intelligence}

\begin{document}

\begin{frontmatter}

\title{Efficient Gesture Recognition for the Assistance of Visually Impaired People using Multi-Head Neural Networks}
% Efficient Gesture Recognition using Multi-Head Neural Networks for the Assistance of Visually Impaired People
% Multi-Head Neural Networks for Efficient Interactive Gesture Recognition in Visually Impaired People Assistance

% Interactive Gesture recognition via Multi-head Neural Network to help Visually Impaired People
% Multi-head neural network-based Interactive Gesture System aimed at helping visually impaired people 
% Assisting visually Impaired people with Efficient Interactive Gesture recognition using Multi-head Neural Networks 
% Efficient Interactive Gesture recognition for assisting visually Impaired people with Multi-head Neural Networks 
% Multi-head Neural Networks for Efficient Interactive Gesture Recognition in Visually Impaired People Assistance
%Interactive Gesture System to Aid People with Visual Impairments}
% Multi-head 
% % TÍTULO ANTERIOR: Real-time Hand Gesture Recognition with CNN

\author[add1]{Samer Alashhab}
\ead{salashhab@ua.es}

\author[add1]{Antonio Javier Gallego\corref{cor1}}
\ead{jgallego@dlsi.ua.es}
\cortext[cor1]{Corresponding author: 
Tel.: +349-65-903772;
Fax: +349-65-909326}

\author[add2]{Miguel Ángel Lozano}
\ead{malozano@ua.es}

\address[add1]{Dept. of Software and Computing Systems, University of Alicante, Carretera San Vicente del Raspeig s/n, Alicante, 03690, Spain}

\address[add2]{Dept. of Computer Science and AI, University of Alicante, Carretera San Vicente del Raspeig s/n, Alicante, 03690, Spain}

%\affil[1]{Dept. of Software and Computing Systems, University of Alicante   \href{mailto:salashhab@ua.es}{salashhab@ua.es} , \href{mailto:jgallego@dlsi.ua.es}{jgallego@dlsi.ua.es}}

%\affil[2]{Dept. of AI and Computational Science, University of Alicante \href{mailto:malozano@ua.es}{malozano@ua.es}}

% \authorrunning{Alashhab et al.} % abbreviated author list (for running head)

%%%%%%%%%%%%%%%%%%%%%%%%%%%%%%%%%%%%%%%%%%%%%%%%%%%%%%%%%%%%%%%%%%%%%%%%%%%%%%%%
\begin{abstract}
This paper proposes an interactive system for mobile devices controlled by hand gestures aimed at helping people with visual impairments. This system allows the user to interact with the device by making simple static and dynamic hand gestures. Each gesture triggers a different action in the system, such as object recognition, scene description or image scaling (e.g., pointing a finger at an object will show a description of it). The system is based on a multi-head neural network architecture, which initially detects and classifies the gestures, and subsequently, depending on the gesture detected, performs a second stage that carries out the corresponding action. This multi-head architecture optimizes the resources required to perform different tasks simultaneously, and takes advantage of the information obtained from an initial backbone to perform different processes in a second stage. To train and evaluate the system, a dataset with about 40k images was manually compiled and labeled including different types of hand gestures, backgrounds (indoors and outdoors), lighting conditions, etc. This dataset contains synthetic gestures (whose objective is to pre-train the system in order to improve the results) and real images captured using different mobile phones. The results obtained and the comparison made with the state of the art show competitive results as regards the different actions performed by the system, such as the accuracy of classification and localization of gestures, or the generation of descriptions for objects and scenes.
\end{abstract}


\end{frontmatter}

%\linenumbers

%%%%%%%%%%%%%%%%%%%%%%%%%%%%%%%%%%%%%%%%%%%%%%%%%%%%%%%%%%%%%%%%%%%%%%%%%%%%%%%%
\section{Introduction}

Gestures are an important part of our communication. They are a form of non-verbal exchange of information that have aroused great interest as regards the design of Human-Computer Interaction (HCI) systems, as they allow users to express themselves naturally and intuitively in different contexts~\cite{hand-gloves}. In some scenarios, a single gesture may be more effective than words (e.g., a \textit{pinch} gesture makes it easier to express the desired zoom level, than explaining it with words).

Hand gesture recognition methods have a significant number of applications, such as controlling unmanned air vehicles (UAVs)~\cite{8101666}, recognizing sign language~\cite{Pigou2015}, or manipulating objects in virtual reality environments~\cite{lin2017vr} or in 3D design tools~\cite{wang2007cad}. In the case of applications such as object manipulation, it is necessary to track the pose of hand and fingers, whereas other applications have to classify the gesture into certain categories, which is the case of sign language recognition. Both dynamic and static gestures are used in these latter applications, depending on whether or not they change over time, respectively~\cite{prakash2019hand}.

One of the contexts in which hand gestures play a prominent role is the field of assistive technologies for people with visual impairments, in which a good user interaction design is of vital importance~\cite{Coughlan2012}. Some devices and applications in this field could greatly benefit from an agile, natural and intuitive interaction system that employs hand gestures. Examples of these devices are OrCam MyEye\footnote{\url{https://www.orcam.com}}, which reads text and identifies objects in the scene, or the eyewear object recognition device proposed by Pintado et al.~\cite{Pintado2019}, which assists people with visual impairments in a market setting. There are also mobile applications such as SuperVision for Cardboard\footnote{\url{http://supervisioncardboard.com}}, which turns a smartphone and a Google Cardboard device into low-cost electronic glasses. However, these systems are limited to a very specific action, requiring the user to press or switch the application to perform another task. A hand gesture-based interface could, therefore, play a key role in improving these technologies.

\begin{figure}[ht]
	\centering
	\begin{subfigure}[t]{0.4\textwidth}
		\centering
		\includegraphics[width=1\textwidth]{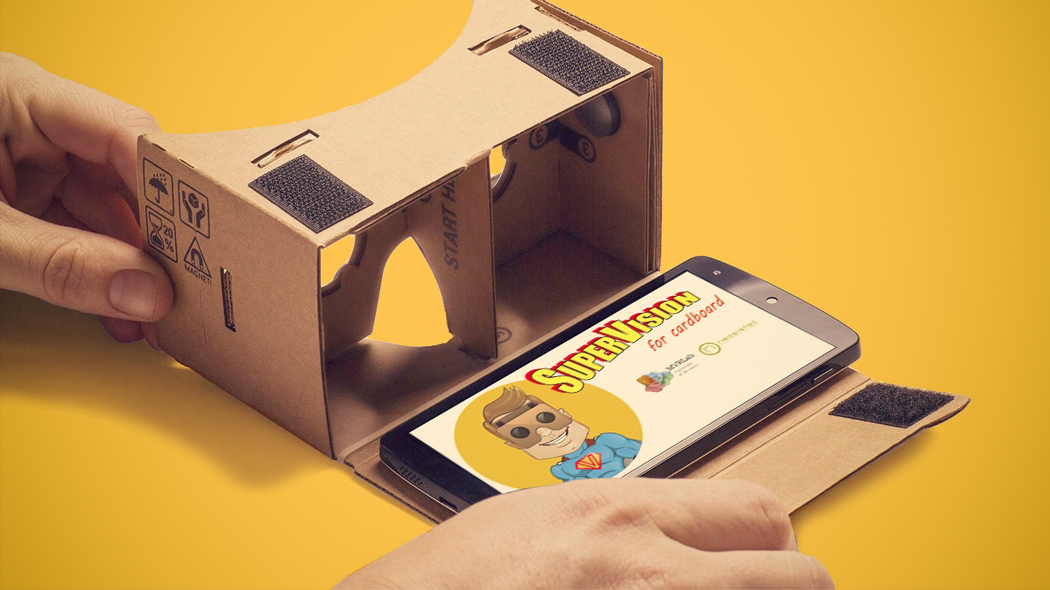}
		\caption{Low cost Google Cardboard. This image was taken from~\cite{Supervision}.}
		\label{fig:cardboard_example_supervision}
	\end{subfigure}
	\quad \quad \quad
	\begin{subfigure}[t]{0.35\textwidth}
		\centering
		\includegraphics[width=1\textwidth]{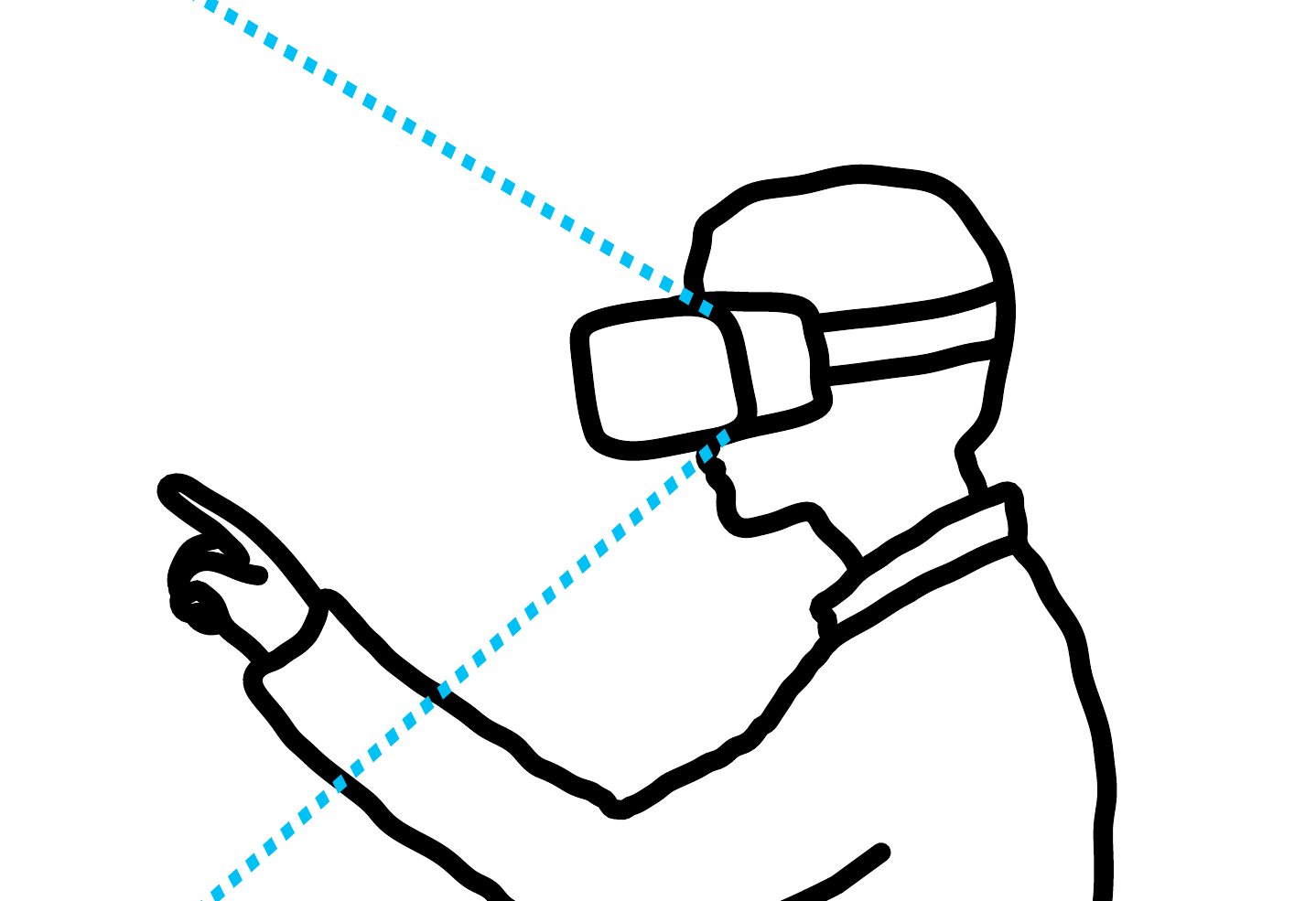}
		\caption{Scheme to illustrate the user interface of the proposed system.} 
		\label{fig:scheme-illustration}
	\end{subfigure}
	\caption{Google Cardboard and scheme of the proposed user interface.}
	\label{fig:ideaIlustration}
\end{figure}

Our goal is to develop a gesture recognition method on which to build an interactive low-cost system for mobile devices controlled by hand gestures (see Figure~\ref{fig:cardboard_example_supervision}), with the objective of helping people with visual impairments. In our proposal, gesture recognition is performed from an egocentric point of view~\cite{tekin2019h}, as the method is intended for applications based on RGB cameras located at the point from which the user views the object (see Figure~\ref{fig:scheme-illustration}). In this respect, gestures could be used to interact with the application, triggering actions such as identifying the object the user is pointing to, describing the scene when the user makes a static loupe gesture, or zooming in and out when a dynamic pinch gesture is employed.

The first step towards this idea was initially developed in \cite{HandGestureDetection2018}, focusing on the classification task and considering a limited set of gestures. In this paper, a completely different and novel approach is proposed to perform multiple tasks simultaneously (including, in addition to classification, actions such as localization and captioning, among others). Also, both the dataset and the experimentation carried out are considerably extended. 

In summary, \textbf{the main contributions made in this work are}:

\begin{itemize}
    \item A novel multi-task architecture that results in a much more efficient and effective model than the use of different separate networks. The proposal is based on a multi-head neural network that integrates the recognition of dynamic and static gestures, object localization and image description functions in the same architecture. Each head of this network is dedicated to a function associated with a given gesture and is executed only if that gesture is detected by the backbone.
 
    \item An exhaustive experimentation of all the parts and actions of the proposed architecture. This experimentation shows that the proposal, in addition to obtaining good results for the different actions carried out, also runs in almost real time on mobile devices.

    \item Our contributions also include the development of a dataset with both real and synthetic images that are annotated at different levels, including gesture category, gesture and fingertip bounding boxes, and object and scene descriptions. 
\end{itemize}

The rest of the paper is organized as follows:  Section~\ref{sec:related_work} shows a review of the state of the art regarding hand gesture, object and image recognition, while the proposed application interface is described in Section~\ref{sec:interface}, the datasets used to train and evaluate our model are detailed in Section~\ref{sec:dataset}, and the proposed approach is introduced in Section~\ref{sec:method}. A comprehensive set of experiments is then shown in Section~\ref{sec:experiments}. Finally, our conclusions and future work are addressed in Section~\ref{sec:conclusions}.

%----------------------------------------------------------------------------
\section{Related work} 
\label{sec:related_work}

Hand gesture recognition approaches include both methods based on dedicated hardware (or other props) and computer vision-based methods~\cite{sonkusare2015review}. The first group contains solutions based on gloves equipped with sensors~\cite{mazumdar2013} and on gloves marked with colors~\cite{wang2009real, lamberti2011real}. In this second approach, each color is used to identify different parts of the hand, which can be detected and tracked by a camera, and no additional hardware is required. In addition to gloves, there are other kinds of sensors whose purpose is also to recognize hand gestures, such as wearable devices that monitor muscle activity on the basis of surface electromyography~\cite{moin2021wearable} or ultrasonic Doppler sensors~\cite{handradar}. 

The second group of hand gesture recognition approaches, which are based on computer vision, has two main categories: appearance-based and 3D model-based methods. The first contains several techniques based on segmenting the hand by color~\cite{pun2011real} (i.e., from RGB images). In many of these approaches, the average radius of the hand is calculated, and the blobs outside that radius are considered to be spread fingers~\cite{perimal2018hand}. The methods in this group have the advantage that they do not need a database of gestures for training~\cite{rajesh2012distance}, but this comes with the limitation that they can recognize only gestures that consist of folded or spread fingers. In~\cite{prakash2017gesture}, the position of the spread fingertips is detected from the vertices of the convex hull of the hand. Other methods rely on depth to segment the hand~\cite{kim2016hand} (i.e., from RGB-D images). There are also similar solutions that rely on the curve of the hand to identify spread fingers~\cite{ren2011depth} and their fingertips~\cite{lai2016fingertips}. 

The group of appearance-based approaches also includes other solutions that use RGB-D images. In~\cite{dinh2014hand}, depth is used to first segment the hand silhouette and remove the background, after which a trained Random Forest (RF) classifier is applied in order to recognize the different parts of the hand in the RGB-D image. Another method~\cite{bamwenda2019recognition} for static gesture recognition obtains a depth-based histogram of oriented gradient features and applies Artificial Neural Networks (ANN) and Support Vector Machines (SVM) for classification. In~\cite{molina2017real}, motion patterns are recognized from sequences of RGB-D images so as to identify dynamic gestures. In order to improve both efficiency and performance when the input is a sequence of images, in~\cite{tang2019fast} key frames are extracted to reduce the number of samples that must be processed. Features are obtained from appearance and motion between consecutive key frames, and a Bag of Features (BoF) approach is applied to classify hand gestures. 

With regard to the general recognition of gestures from RGB and RGB-D images, the methods that have been shown to be most effective are those based on Deep Neural Networks (DNN). Most of them use Convolutional Neural Networks (CNN), which have obtained excellent results for image recognition~\cite{schmidhuber2015deep}. Architectures used for this purpose are those such as AlexNet~\cite{Krizhevsky2012}, GoogleNet~\cite{Szegedy2015b}, DenseNet~\cite{Huang2017densenet}, ResNet~\cite{Kaiming2016resnet}, Xception~\cite{Chollet2016XceptionDL}, and lightweight architectures such as MobileNet~\cite{howard2017mobilenets}, SqueezeNet~\cite{iandola2016squeezenet} and EfficientNet~\cite{Tan2019EfficientNet}, on which several works obtaining excellent results have been based~\cite{HandGestureDetection2018}. With regard to gesture classification, there are also CNN-based approaches such as~\cite{Lin2014}, which embodies a previous image calibration step. CNNs have also been applied in order to recognize sign language in a single frame~\cite{Bheda2017} or a sequence of frames~\cite{Pigou2015} (dynamic gestures). In~\cite{Molchanov2015}, the CNN takes both intensity and depth video sequences as input for the recognition of dynamic gestures with the objective of designing touchless interfaces in cars. In addition to CNNs, other deep learning-based approaches are also used to segment hands by depth, as is the case of SegNet, a deep convolutional encoder-decoder architecture proposed to detect fingertips~\cite{nguyen2019hand}. Dynamic gesture recognition in video sequences has also been addressed by means of Bidirectional Long Short-Term Memory (BiLSTM)~\cite{xie2017deep} and Temporal Segment Networks (TSN)~\cite{s21020356} in order to capture temporal information. 

Most approaches based on 3D models use the skeleton of the hand. However, there are also volumetric solutions~\cite{ge20193d}. The skeleton can be obtained using specific hardware, such as instrumented gloves, although there are also devices based on depth cameras that provide a built-in joint tracker, such as Microsoft Kinect~\cite{xi2018real} or Leap Motion Controller\footnote{\url{https://www.ultraleap.com/product/leap-motion-controller/}}, which has been applied to, for example, hand function rehabilitation~\cite{xiao2021research}. There are also libraries, such as OpenPose hand detection\footnote{\url{https://github.com/CMU-Perceptual-Computing-Lab/openpose}}~\cite{8765346, simon2017hand} or MediaPipe Hands\footnote{\url{https://mediapipe.dev}}~\cite{zhang2020mediapipe}, that make it possible to obtain the skeleton in 2D or 3D from an RGB image. There are also some methods that recover the 3D skeleton of a hand from RGB-D images~\cite{ge20193d} or from a sequence of raw RGB images~\cite{tekin2019h}. This 3D data information of the hand skeleton (joint coordinates) has been used as a basis for different types of classifiers with which to recognize gestures, such as Hidden Markov Models (HMM) and Dynamic Time Warping (DTW) in~\cite{raheja2015robust} or CNNs in~\cite{devineau2018deep}. A comparison~\cite{de20173d} between an SVM classifier using skeleton data and a CNN classifier using RGB-D images shows that the former approach provides superior results as regards hand gesture recognition. However, the use of libraries such as OpenPose or MediaPipe to obtain skeleton data from RGB images has certain limitations, since it is not possible to recognize accurately skeleton points for several poses~\cite{9616851}.

Our work focuses on the development of a low-cost general gesture recognition solution that could be integrated into most of the current smartphones equipped with RGB cameras. For this, we rely on deep neural networks that take an RGB image as input. Besides gesture recognition, our purpose is to integrate the functions required to identify objects, describe the scene or zoom in and out into the same network. In order to do this efficiently, we have followed an approach similar to that of~\cite{kopuklu2019real}, in which a two-stage network architecture is implemented to build a real-time gesture recognizer: the first stage is dedicated to the detection of the gesture, and the second one to the classification, which is executed only if the presence of a gesture is detected in the image. In our case, we propose a multi-head architecture with a backbone dedicated to the classification of gestures, and a set of specialized heads for each gesture, which will be triggered only if the corresponding gesture is detected. 

Depending on the gesture detected, a given action is, therefore, performed using a specialized head: object recognition, image description and zoom in/out. We considered various state-of-the-art DNN architectures for the object recognition head, such as You Only Look Once (YOLO)~\cite{redmon2018yolov3}, Faster R-CNN (FRCNN)~\cite{ren2015:fasterrcnn}, and RetinaNet~\cite{Lin2017RetinaNet}. These models are able to identify multiple objects in the image and their bounding boxes. We also compared a modified version of the Filter Selection (FS)~\cite{filterselection} approach as an alternative to these object recognition methods, in which a set of filters from the backbone is selected in order to calculate the location of the gesture in the image. This method has the advantage of not adding extra modules to the original architecture. With regard to the head employed to obtain the description of the image, it can be addressed using image captioning methods~\cite{hossain2018captioningsurvey}. For this, we have also considered different models~\cite{you2016image, tanti2018put}, which usually combine a CNN in order to extract features from the image, and a Recurrent Neural Network (RNN) to generate the description. Finally, with regard to zooming in and out with the pinch gesture, we propose our own architecture as a specialized head for this task.

% ---------------------------------------------------------------------------
\section{Application interface} 
\label{sec:interface}

One of the areas in which improvements could be made to the applications aimed at helping people with visual impairments, such as \textit{Supervision for Cardboard}~\cite{Supervision}, is the interface. An important enhancement to its usability would be that users could use the application while moving or performing other tasks, without having to touch the mobile screen. 

To this end, this paper proposes an interactive system for mobile devices controlled by hand gestures. Users could install their mobile phones on Virtual Reality Glasses (VRG) or on a low cost Google Cardboard (see Figure \ref{fig:cardboard_example_supervision}), and view the environment directly through the mobile screen. The proposed system would allow them to interact with the device using different hand gestures and would use augmented reality to display the result of the actions on the screen (see Figure \ref{fig:scheme-illustration}).

\begin{figure}[ht]
	\centering
	\begin{subfigure}[t]{0.18\textwidth}
		\centering
		\includegraphics[width=1\textwidth]{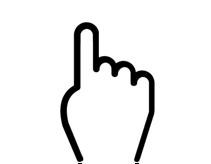}
		\caption{Point}
		\label{fig:gesture-point}
	\end{subfigure}
	\begin{subfigure}[t]{0.18\textwidth}
		\centering
		\includegraphics[width=1\textwidth]{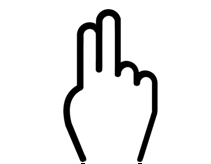}
		\caption{Drag}
		\label{fig:gesture-drag}
	\end{subfigure}
	\begin{subfigure}[t]{0.18\textwidth}
		\centering
		\includegraphics[width=1\textwidth]{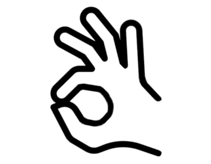}
		\caption{Loupe}
		\label{fig:gesture-loupe}
	\end{subfigure}
	\begin{subfigure}[t]{0.18\textwidth}
		\centering
		\includegraphics[width=1\textwidth]{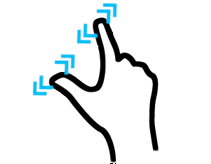}
		\caption{Pinch}
		\label{fig:gesture-pinch}
	\end{subfigure}
	\newline
	\vspace{0.5cm}
	\newline
	\begin{subfigure}[t]{0.18\textwidth}
		\centering
		\includegraphics[width=1\textwidth]{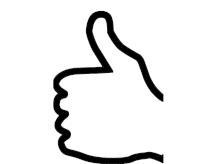}
		\caption{Thumb-up}
		\label{fig:gesture-thump-up}
	\end{subfigure}
	\begin{subfigure}[t]{0.18\textwidth}
		\centering
		\includegraphics[width=1\textwidth]{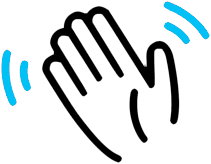}
		\caption{Wave}
		\label{fig:gesture-wave}
	\end{subfigure}
	\caption{Graphical description of the proposed hand gestures. The gestures used to interact with the interface are Point, Drag, Loupe and Pinch. The system will consider Thumb-up and Wave as \textit{other gestures}. Point, Drag, Loupe, and Thumb-up are static gestures, while Pinch and Wave are dynamic gestures.}
\label{fig:gestures}
\end{figure}

A set of four simple gestures is proposed as a user interface to interact with the system: point, drag, loupe and pinch (see Figures \ref{fig:gesture-point}--\ref{fig:gesture-pinch}). There are three static gestures (point, drag and loupe) with which to execute specific actions, and a dynamic gesture (pinch) with which to zoom-in and zoom-out the image. A better description of the four gestures proposed is provided below:

\begin{itemize}
\item \textit{Point}: Static gesture formed by extending the index finger and flexing the remaining fingers into the palm. This gesture allows users to point to the objects of which they wish to obtain a description.
\item \textit{Drag}: Static gesture formed by pointing with both the index and the middle fingers. This gesture allows users to freeze the image while simultaneously performing a panning movement of the scene following their fingertips. This is useful in combination with the zoom gesture. 
\item \textit{Loupe}: Static gesture formed by joining the thumb and the index finger to form the shape of a circle, and leaving the remaining fingers extended. This gesture shows more information about the scene and the objects that appear in it.
\item \textit{Pinch}: Dynamic gesture formed by moving the thumb and the index finger toward each other or away from each other, in order to perform a zoom-in or a zoom-out operation, respectively. It is equivalent to the pinch gesture used on touch screens. This dynamic gesture allows the zoom level to be controlled with the movement of the fingers.
\end{itemize}

This small set of gestures has been selected in order to allow an intuitive and easy interaction with the system. However, our proposal is designed in a generic manner (as will be shown in Section \ref{sec:method}), signifying that it would be easy to add new gestures so as to expand its functionality, if necessary.

It is important to note that, in addition to the proposed gestures, the system has to identify whether or not there is a gesture present in the image, and it has to differentiate these gestures from any other possible gestures, both static and dynamic (such as the thumb-up and wave gestures shown in Figure \ref{fig:gestures}).

% ------------------------------------------------------
\section{Datasets}
\label{sec:dataset}

Three datasets were created in order to train and evaluate the proposed model\footnote{These datasets are freely available for the scientific community on demand at \url{https://www.dlsi.ua.es/~jgallego/datasets/gestures}}: a dataset containing real images of hand gestures, a synthetic dataset used to pre-train the system, and a dataset containing descriptions of scenes used by the captioning actions.

\begin{figure}[ht]
	\centering
\includegraphics[width=1\textwidth]{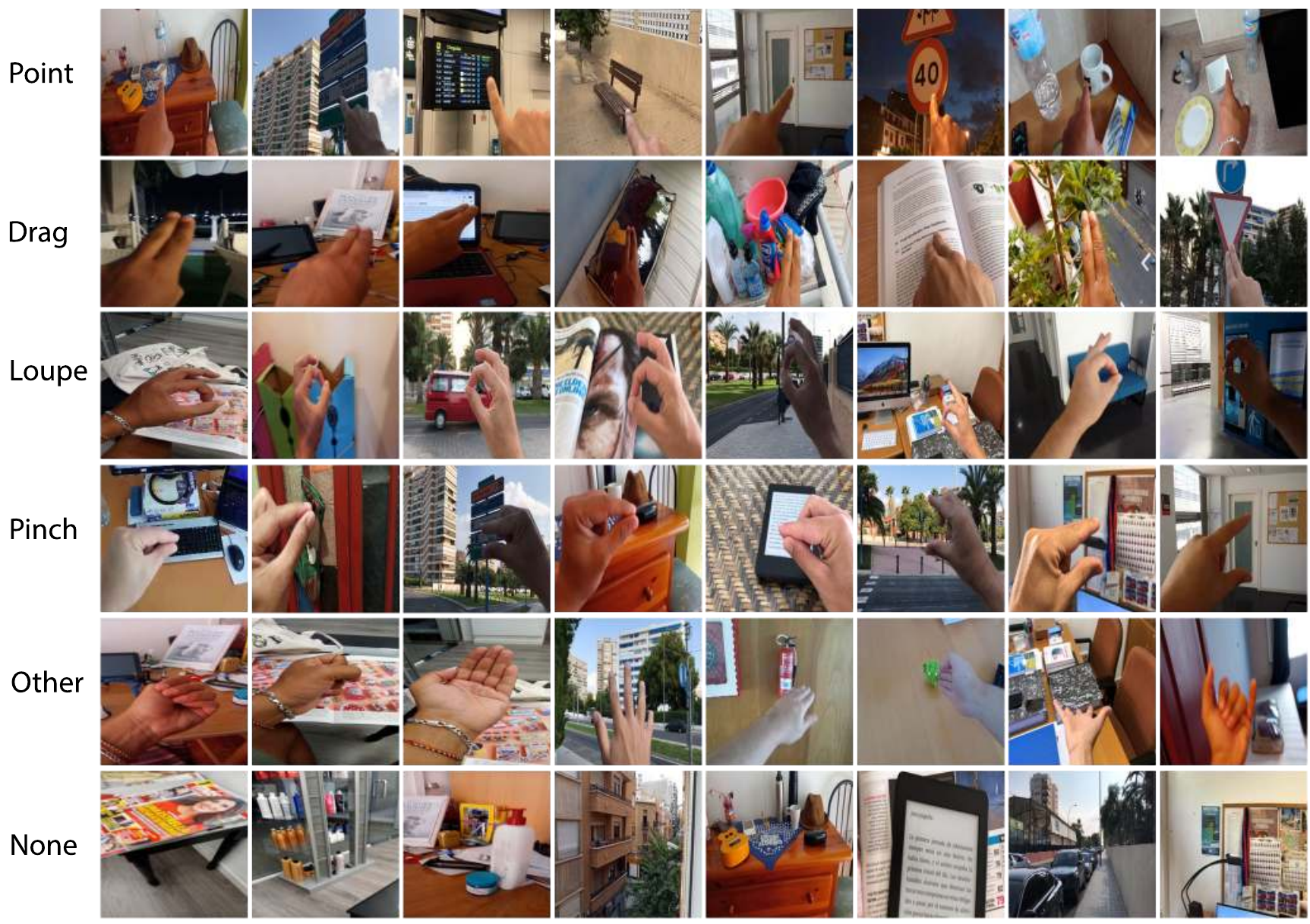}
\caption{Some samples of the dataset with real images. The first four rows show the different gestures proposed in order to interact with the interface. The last two rows include some examples of the ``\textit{other gestures}'' and the ``\textit{no gestures}'' classes.} 
\label{fig:dbexamples}
\end{figure}

The samples of the dataset with real images were extracted from videos obtained with mobile phones. In order to ensure that the data generated were varied, different phone cameras were used to record indoor and outdoor scenes, with varied backgrounds and under different lighting conditions. In addition to images of the four proposed gestures, samples of backgrounds without gestures and of gestures other than those proposed were also extracted from these videos. The background images were used to train the system to discriminate between the presence or absence of hands, and the ``\textit{other-gestures}'' images were employed to assess whether the system was capable of differentiating them from the proposed gestures. Note that in the case of the dynamic pinch gesture, an average of 10 consecutive frames was extracted for each gesture in order to train and evaluate the methods used to detect the movement of the fingers and their position. A total of 13,559 frames were extracted from the original videos, trying to balance the number of samples selected for each class. Figure~\ref{fig:dbexamples} shows some examples of the images included in this dataset. 

\begin{figure}[ht]
	\centering
\includegraphics[width=.7\textwidth]{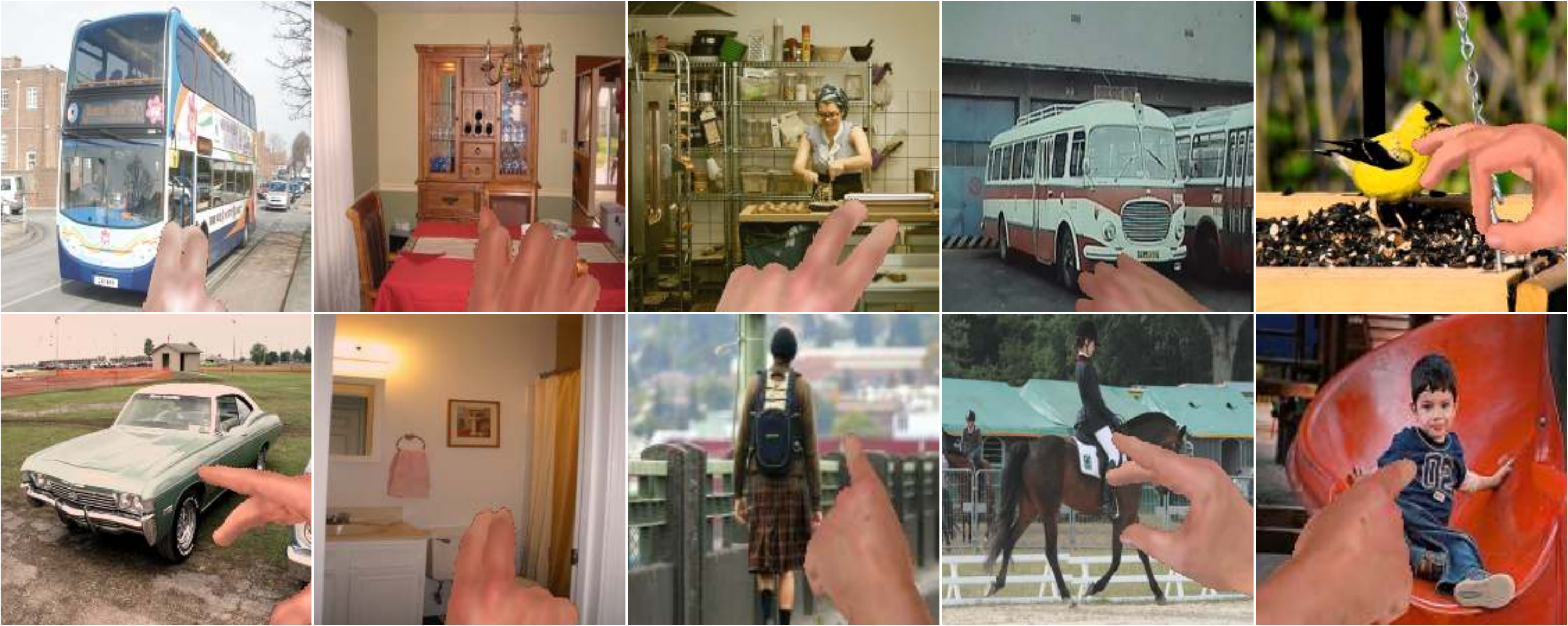}
\caption{Some examples of the images generated for the synthetic dataset.} 
\label{fig:synth_examples}
\end{figure}

In order to improve the results obtained and help the training process, a synthetic dataset was created using a modified version\footnote{https://github.com/malozano/libhand} of the LibHand tool~\cite{libhand}, an open-source library for the rendering of human hand poses. We modified this library so as to enable the definition of gestures through a set of rules with the ranges of movement allowed for the finger joints, thus enabling variations of these gestures to be generated randomly within these ranges. This tool was used in order to automatically generate and label a dataset with a total of 13,200 images (2,200 of each gesture), with random variations in position, in the shape of the gestures, in the color of the skin, including random blur to emulate the motion effect, and with different backgrounds (using random images from Flickr8k \cite{flicker8k} and Visual Object Classes (VOC) \cite{everingham2015pascal} datasets). Figure \ref{fig:synth_examples} shows some examples of the gestures generated.

These two datasets were labeled, indicating both the category and the position of the gesture. For the position, the coordinates were annotated using bounding boxes for 1) the position of the hand within the image, 2) the position of the fingertips, and 3) the coordinates and labels of the objects pointed to.

As explained in the previous section, the loupe gesture triggers the action of displaying a description of the scene that appears in the image. In order to train and evaluate systems capable of generating these descriptions, it was necessary to use an additional dataset of images with the corresponding associated descriptions. For this, we used the Flickr8k dataset \cite{flicker8k}, which contains 8,000 images manually selected from the Flickr website with five descriptions of each image. Besides, we added a subset of 4,000 images from our dataset of real images, which also included 5 descriptions per image (this allowed the proposed system to adapt to our type of data, i.e., images of gestures taken with mobile phones). This subset includes both loupe and point gesture images (2,000 for each). The point gesture was included in order to increase the variability, and also make it possible to use the captioning head for these gestures (which could be appropriate for some applications).  

The original resolution used for videos and images was 1,920$\times$1,080 pixels. However, after conducting a series of initial performance and accuracy experiments at different resolutions, and also motivated by the restrictions of some of the methods evaluated, we decided to scale the images to a spacial resolution of 224$\times$224 pixels. Table \ref{tab:dataset-table} shows a summary of the three datasets considered in this work, including the number of samples per class in each dataset.

\begin{table}[ht]
\centering
\small
\caption{Summary of the datasets, including class name, number of samples per class, and a short description of the content of each class.}
\label{tab:dataset-table}
%\begin{tabular}{ccl}
\begin{tabularx}{\linewidth}{lcccX}
\toprule
\textbf{Gesture} 
& \begin{tabular}[c]{@{}c@{}}\textbf{\# Synthetic}\\\textbf{samples}\end{tabular} 
& \begin{tabular}[c]{@{}c@{}}\textbf{\# Real}\\\textbf{samples}\end{tabular} 
& \begin{tabular}[c]{@{}c@{}}\textbf{\# Captioning}\\\textbf{samples}\end{tabular} 
& \textbf{Description} \\ 
\midrule
Point            & 2,200    & 2,088     & 2,000     & Images of pointing gestures. \\
Drag             & 2,200    & 2,143     & --        & Images including drag gestures.  \\
Loupe            & 2,200    & 2,147     & 2,000     & Samples including loupe gestures. \\
Pinch            & 2,200    & 2,066     & --        & Sequences of dynamic pinch gestures.  \\
Other            & 2,200    & 2,121     & --        & Images of gestures other than the four defined. \\
None             & 2,200    & 2,994     & 8,000     & Samples in which no hand appears. \\ 
\midrule
Total            & 13,200   & 13,559    & 12,000   & Grand total: 38,759 \\
\bottomrule
%\end{tabular}
\end{tabularx}
\end{table}

%%%%%%%%%%%%%%%%%%%%%%%%%%%%%%%%%%%%%%%%%%%%%%%%%%%%%%%%%%%%%%%%%%%%%%%%%%%%%%%%
\section{Method} 
\label{sec:method}

\begin{figure}[ht]
\centering  
\includegraphics[width=.85\linewidth]{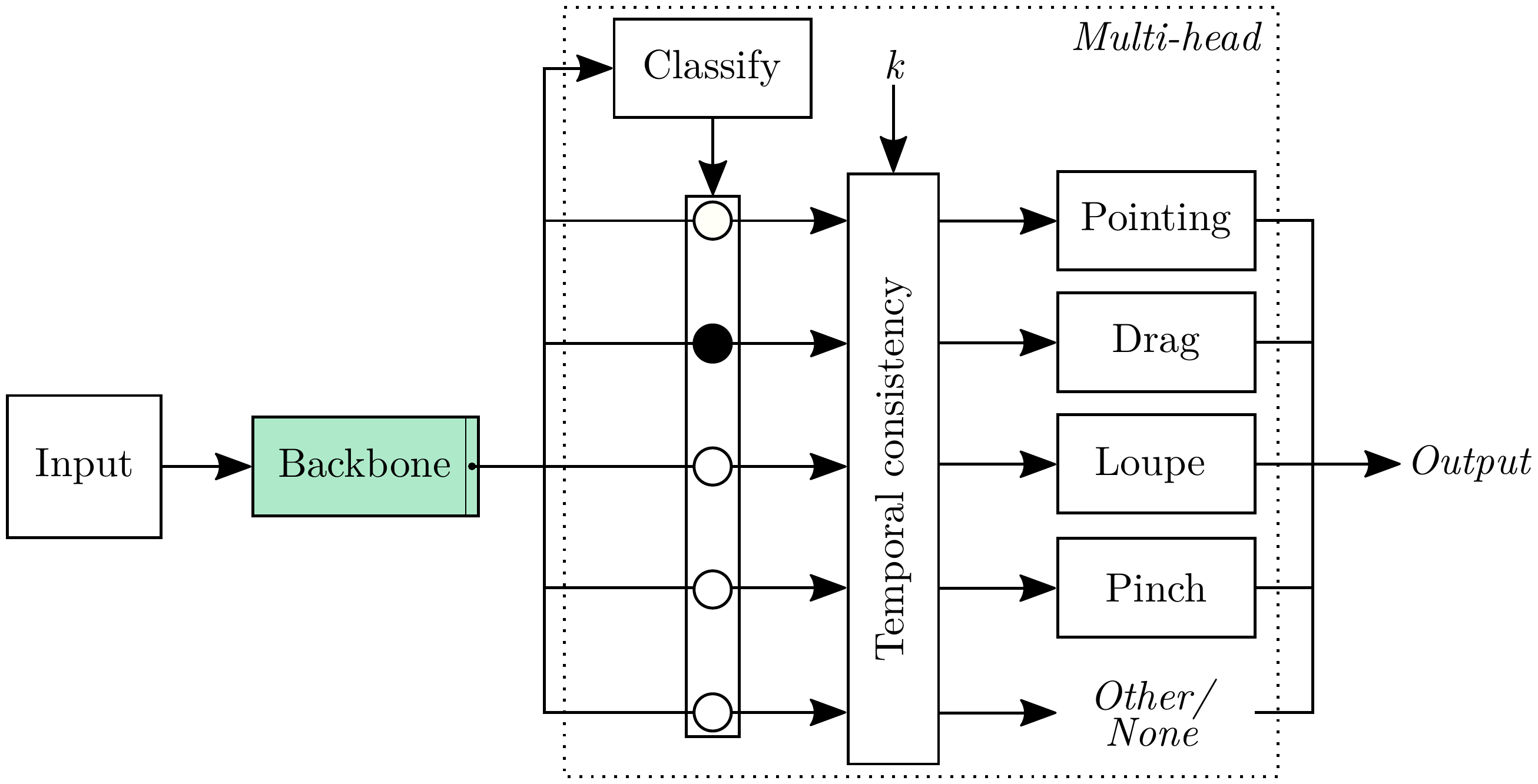}
\caption{Scheme of the proposed two-step multi-head network.}
\label{fig:scheme}
\end{figure}

The input received by the system is a sequence of frames captured with a mobile phone camera. The proposed approach processes each of these frames in order to first classify the gesture that appears in the image and then perform an action based on the gesture detected. This is done using an architecture divided into two stages (see Figure \ref{fig:scheme}): 1) an initial backbone processes the image in order to extract a set of representative features, and 2) these features are then used to classify the gesture and perform an action by means of the head specialized in the gesture detected. 

In the second step, the common features extracted by the backbone are first processed using the ``Classify'' head shown in Figure \ref{fig:scheme}, which yields an $L$-dimensional one-hot vector, where $L$ is the number of possible gestures. The other specialized heads are activated or deactivated through the use of a switch-type layer that queries the value set to one in this one-hot vector.

The main advantage of this architecture is that it can carry out multiple processes with a reduced number of parameters and, therefore, with fewer hardware requirements. To achieve this, the same initial features are used for all the actions to be carried out. Moreover, in the second stage of the method, only one of the specialized heads is activated depending on the gesture detected, which also improves performance. 

The stability of the system has been improved by establishing a time margin of $k$ frames with which to validate the detection of a gesture. That is, the system processes the frames of the input video and returns a response for each one. However, in order to ensure a consistent response for consecutive frames, at least $k$ frames must maintain the same response to consider the prediction valid. In the tests carried out, it was sufficient to establish this time margin as 3 frames, as will be shown in the experimentation section. 

As Figure \ref{fig:scheme} shows, this temporal consistency applies to all gestures, including the ``Other'' and ``None'' categories, which are joined and considered as a single negative class and do not perform any specialized processing, but simply do not return a response. In this way, when the (positive) gestures considered are no longer detected during $k$ frames, no response will, therefore, be provided. 

The following subsections provide detailed descriptions of the different parts of this approach.

%---------------------------------------
\subsection{Backbone and classification}
\label{sec:classification}

The first step in the method processes each input frame using a backbone to obtain a common feature vector that is then used to carry out the remaining actions. This part of the method is the most important, since the efficiency and efficacy of the proposed solution depends on its result. It was for this reason that a total of 18 different approaches were compared, including network architectures such as MobileNet~\cite{howard2017mobilenets}, EfficientNet~\cite{Tan2019EfficientNet}, Xception~\cite{Chollet2016XceptionDL}, and SqueezeNet~\cite{iandola2016squeezenet}. The results obtained will be analyzed in detail in Section \ref{sec:classification_results}. However, we anticipate that the approach that obtained the best results (when considering the balance between precision and execution time) was Darknet-53~\cite{redmon2018yolov3}. 

Darknet-53 is the backbone used by YOLO v3. It is made up of 53 convolutional layers with residuals or shortcut connections (the complete architecture can be found in \cite{redmon2018yolov3}, see Table 1). This network is more efficient and obtains better results than its previous versions or other similar architectures. The last convolutional layer of this backbone is connected with a Global Average Pooling (GAP) operation and then with the other layers of the module or the head in question (see Figure \ref{fig:scheme}). 

With regard to the classification head, it is necessary only to add a dense layer with $L$ neurons and with the SoftMax activation function in order to fine-tune to the $L$ classes in our dataset, as is usual in transfer learning tasks~\cite{ChatfieldSVZ14}. As indicated previously, an $L$-dimensional one-hot vector is obtained as a result of this classification, in which the gesture detected is marked as 1 and the others as 0. This result is used to activate only the head corresponding to the gesture detected. 

The proposed methodology has the additional advantage of allowing new gestures to be easily added. This can be done by adding the new category to the classification head and then fine-tuning the backbone in order to detect the new gesture.

%---------------------------------------
\subsection{Pointing and drag gestures}
\label{sec:fs}

The actions corresponding to these two gestures share a common first part: the detection of the tip of the extended fingers. Up to 8 possible approaches were compared for this process, including object detection methods such as YOLO~\cite{redmon2018yolov3}, RetinaNet~\cite{Lin2017RetinaNet}, and Faster R-CNN~\cite{ren2015:fasterrcnn} (as will be seen in Section~\ref{sec:pointing_evaluation}). It was eventually determined that the approach that obtained the best results as regards both precision and efficiency was an approximation based on the Filter Selection (FS) method~\cite{filterselection}. 

%---------------------------------
\subsubsection{Filter Selection}

FS is a weakly-supervised object detection method that selects the set of filters from a categorical CNN that maximizes the detection precision of the classes considered. That is, this method does not add extra layers, but directly takes advantage of the filters already learned for the classification task and reuses them to obtain the location of the objects for each class. It is, therefore, also more efficient, as it does not require extra parameters. This solution, which was initially proposed for other types of tasks, has been modified to support multiple classes and to speed up the detection process. 

FS analyzes all the filters learned by the categorical network (denoted as ${\cal F}$) in order to then select a subset ${\cal F}^c \subseteq {\cal F}$, which will be used to determine the location of the class $c$. This is done by calculating the Intersection over Union (\text{IoU}) between the ground-truth and the predictions $P_{f}^{(i)}$ for the filter $f$ and the image $i$ of the set of images $I^c$ with samples of the searched class $c$. Only those filters whose average IoU is greater than a threshold $\alpha$ are selected from this result. The subset of filters ${\cal F}^c$ is formally calculated as follows: 

\begin{equation}
{\cal F}^c = \left\{ 
	f \in {\cal F} \mid 
	\frac{1}{|I^c|} \sum_{i=1}^{|I^c|} \textit{IoU}(P_{f}^{(i)}, B_{c}^{(i)}) > \alpha 
\right\}
\label{eq:Fc}
\end{equation}

\noindent 
where $B_{c}^{(i)}$ are the ground-truth localizations for the image $i$ and class $c$, and $|I^c|$ represents the cardinality of the set $I^c$. The prediction set $P_{f}^{(i)}$ for an input image $i$ and a filter $f$, is computed as follows: 

\begin{equation}
P_{f}^{(i)} = \textit{Blobs}((r(A_{f}^{(i)}) > \beta) \oplus s)
\label{eq:P}
\end{equation}

\noindent 
where $A_{f}^{(i)}$ represents the activation map (also known as the feature map) obtained for the filter $f$ and the input image $i$. Unlike that which occurs in the original method, our approach does not perform a backpropagation pass, but rather uses the activation maps directly, thus significantly speeding up the entire process. The activations obtained are then rescaled to range $[0, 1]$ using the function $r$ and are thresholded using $\beta$ to obtain a binary matrix $\mathbb{R}^{(w \times h)} \rightarrow [0,1]^{(w \times h)}$ that is the same size as the input image, where $w$ and $h$ are the width and height, respectively. A morphological dilation operation (denoted by $\oplus$) is applied using a structuring element $s$. Since the noise is removed by the thresholding operation, the objective of this dilation is to close small gaps and slightly increase the size of the detections. Finally, the function \textit{Blobs} calculates the groups of connected pixels, returning a list of bounding boxes for the blobs detected. 

In Equation \ref{eq:Fc}, in order to calculate the \text{IoU} of the predictions obtained for an input image $i$, each predicted bounding box from the set $P_{f}^{(i)}$ is mapped onto the ground truth bounding box $B_{c}^{(i)}$ with which it had a maximum \text{IoU} overlap (considering that both $P_f$ and $B_c$ may contain many bounding boxes): 

\begin{equation}
	\textit{IoU}(P_{f}^{(i)}, B_{c}^{(i)}) = \frac{\text{area}(P_{f}^{(i)} \cap B_{c}^{(i)})}{\text{area}(P_{f}^{(i)} \cup B_{c}^{(i)})}
	\label{eq:iou}
\end{equation}

Once this stage has been completed, the subset of filters ${\cal F}_c$ for each target class $c$ is stored to be used in the inference stage for unseen images. In our case, this selection process is carried out on the categorical network described in the previous section (i.e., Darknet-53 + classification head), initialized with the pre-trained weights obtained with the ILSVRC dataset~\cite{Krizhevsky2012}, a generic purpose database for object classification, and fine-tuned in order to classify the classes of our dataset (this training process will be explained in Section \ref{sec:train}). The influence of the different parameters of the proposed method on this and on the other network architectures considered will be evaluated in Section \ref{sec:pointing_evaluation}.

%---------------------------------
\subsubsection{Fingertip detection}

In the inference stage, an input sample is forwarded through the trained model (the backbone in Figure \ref{fig:scheme}), and if a pointing or drag gesture is detected, the activation maps of the network are used to obtain its localization. This is done in a similar way to Equation \ref{eq:P}, but by performing the average of the activations obtained from the selected subset of filters ${\cal F}^c$. The function $\textit{FS}(i, c)$ calculates the localization of targets using the pre-calculated subset of filters ${\cal F}^c$, as follows:

\begin{equation}
\textit{FS}(i, c) = \textit{Blobs} \biggl( \biggl( \biggl( \frac{1}{|{\cal F}^c|} 
	\sum_{f \in {\cal F}^c} r(A_{f}^{(i)}) \biggr) > \beta
\biggr) \oplus s \biggr)
\label{eq:FS}
\end{equation}

With regard to the drag gesture, the system needs to know only the position of the tip of the spread fingers, since, for the action to be carried out, this information is sufficient to calculate the movement made by the fingers between consecutive frames. However, in the case of the pointing gesture, it is also necessary to identify and describe the closest object to the fingertip. Depending on the final application, this could be done by means of the captioning generation method detailed in the following section or by applying FS to the categories of the objects to be identified by our system. This would, therefore, allow the system to indicate the class of the object pointed to by consulting the annotation of the ILSVRC dataset~\cite{Krizhevsky2012}. 
%Our dataset also includes annotations for the objects pointed to, so it could be used to evaluate this task.

%---------------------------------------
\subsection{Loupe gesture}

\begin{figure}[ht]
\centering  
\includegraphics[width=1\linewidth]{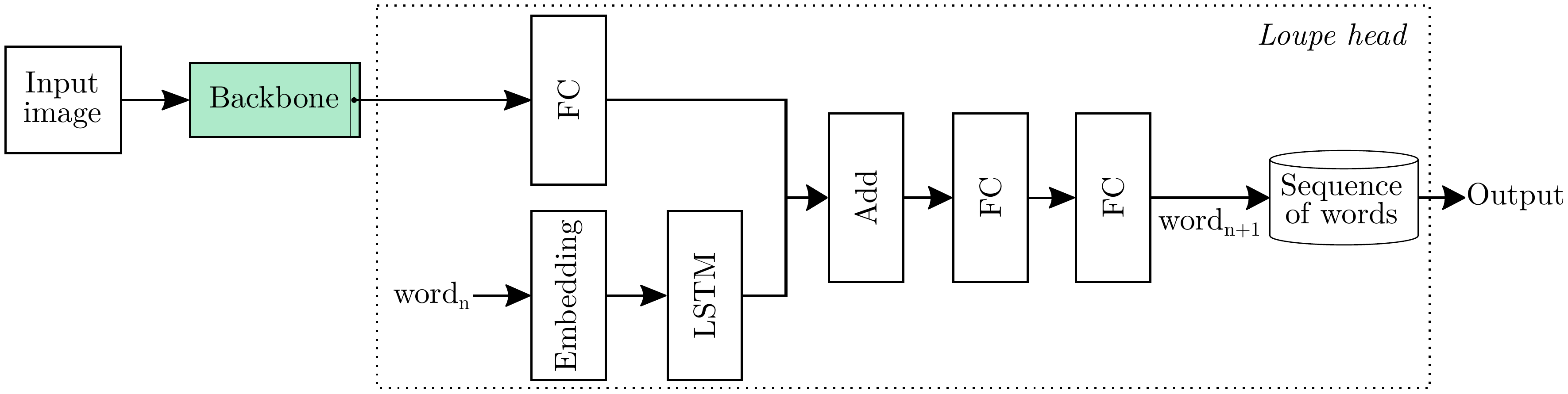}
\caption{Image captioning model implemented by the head that processes the loupe gesture.}
\label{fig:scheme_captioning}
\end{figure}

The objective of the loupe gesture is to obtain a textual description of a scene. A specialized head based on the caption generation model proposed by Marc Tanti et al.~\cite{tanti2018put}, which is known as \textit{merge-model}, was used for this action. This multimodal architecture (see Figure \ref{fig:scheme_captioning}) performs a late fusion of information. As the authors of the original model indicate, results suggest that the visual and linguistic modalities for caption generation need not be jointly encoded by the RNN, as this yields large memory-intensive models with few tangible advantages in performance; the multimodal integration should rather be delayed to a subsequent stage. Late fusion, therefore, makes it possible to use specialized architectures for each modality, such as, in our case, a backbone with which to process the images and an RNN for the text. 

In our implementation, the features obtained by the backbone for the input image are processed through the use of a fully connected (FC) layer with 256 neurons. The output of this layer is then combined with that obtained from the recurrent part of the network used for the linguistic information. This other part of the network is made up of an embedding layer followed by an LSTM layer with 256 neurons. Once the combined features have been obtained, two FC layers, each of which contains 256 neurons, are used to obtain the final result. The ReLU activation function is used in all the layers, with the exception of the last layer, which uses a Softmax activation function to determine the next word predicted by the architecture. During inference, the start token ``\textit{startseq}'' is passed, generating one word, after which the model is recursively called, using the words generated as input, until the end token ``\textit{endseq}'' is obtained or the maximum description length is reached.

Finally, it should also be noted that a post-processing step is eventually carried out on the sentences generated, since the network sometimes generates texts that begin with ``A hand/finger is pointing to...'' or ``A hand and...''. A set of basic rules have, therefore, been defined that modify the phrase in order to remove these texts and thus generate a sentence that refers only to the scene.

%---------------------------------------
\subsection{Pinch gesture}
\label{sec:pinch}

\begin{figure}[ht]
	\centering
\includegraphics[width=.9\textwidth]{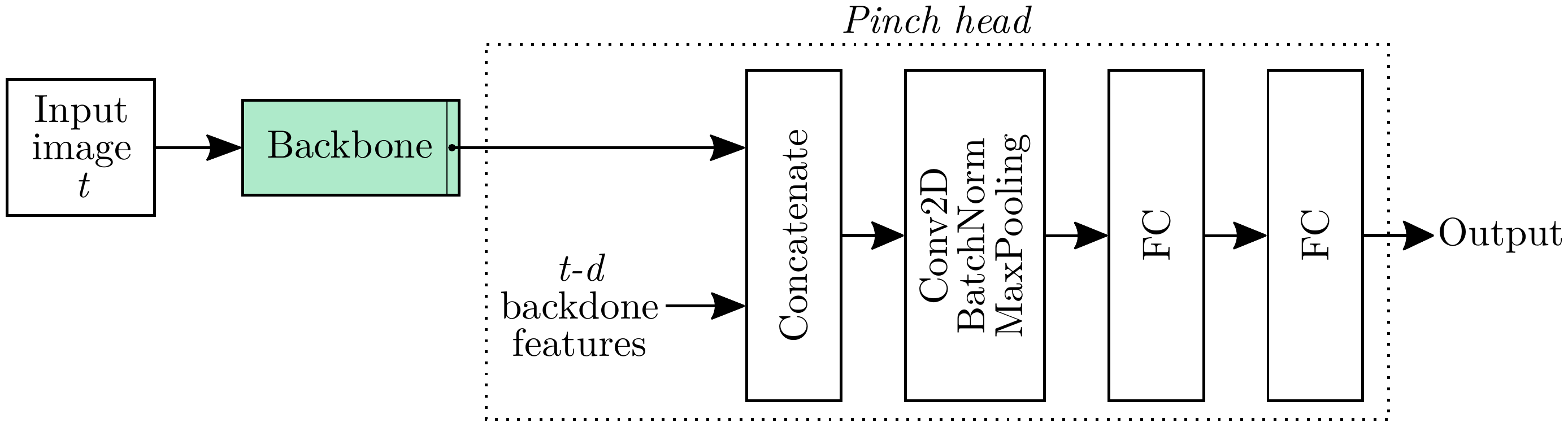}
\caption{Scheme of the architecture proposed for the head that processes the pinch gesture.}
\label{fig:pinch_model}
\end{figure}

The purpose of this gesture is to control the zoom level. When it is detected, the image freezes and the user can increase and decrease the zoom level with the movement of the fingers. The specialized head shown in Figure~\ref{fig:pinch_model} is used to control this action. This head receives two inputs, one containing the features extracted by the backbone for the current frame $t$, and other containing the features obtained for the frame $t-d$. In other words, the frame obtained $d$ previous frames is used rather than the immediately previous one. In our case, after a series of preliminary experiments, we established $d=5$ in order to have a notable difference between the frames compared that facilitates the detection of movement. The history of stored frames is managed internally by the specialized head itself, for which it is sufficient to store a buffer with the last $d$ elements. 

The architecture of this head is very simple (see Figure \ref{fig:pinch_model}), which results in an efficient and effective model. The features of the two inputs (current frame and frame $t-d$) are concatenated and processed by a convolution layer with 64 filters of size 3$\times$3. A Batch Normalization layer is then added, followed by a Max Pooling operation, which is used to reduce dimensionality. The result obtained is connected with an FC layer (containing 32 neurons and a ReLU activation function) through the use of a flatten operation and, finally, with another FC layer with Softmax activation in order to determine whether zoom-in, zoom-out, or no zoom is being performed.

% -------------------------------------------------------
\subsection{Training stage}
\label{sec:train}

Rather than training the entire architecture in one stage using a combined loss function, a two-phase training process is proposed. In the first phase, the backbone is trained for the classification task, while in the second phase, the weights obtained for the backbone and the classification head are frozen and the remaining heads are trained. 

The backbone is initialized using the pre-trained weights obtained with the ILSVRC dataset~\cite{Krizhevsky2012}. Weight initialization is a common practice that makes it possible to obtain better results in less training time~\cite{Yosinski2014}. A fine-tuning process in then applied to the backbone connected with the classification head. This is done using the datasets described in Section~\ref{sec:dataset}: the synthetic dataset is first employed, after which, and based on the weights obtained, a second fine-tuning is performed using the dataset with real images.

The categorical \emph{cross-entropy} loss function between each output activation and its expected activation was used to calculate the error. The network parameters were tuned by means of back-propagation using stochastic gradient descent~\cite{Bottou2010} and considering the adaptive learning rate proposed by \cite{Zeiler2012}. The training was performed a maximum of 200 epochs for each dataset (the synthetic and the real one) with a mini-batch size of 32 samples, and \emph{early stopping} when the loss did not decrease during 10 epochs.

Data augmentation was also used to artificially increase the size of the training set by randomly applying different types of transformations to the original training samples. This technique usually improves the performance and helps reduce overfitting~\cite{Krizhevsky2012, Chatfield2014}. In our case, 10 augmented images were generated for each image in the training set. The transformations applied were randomly selected from the following set of possible transformations: horizontal flips (allowing the system to work regardless of the hand used), horizontal and vertical shifts ([-10, 10]\% of the image size), zoom ([-10, 10]\% of the original image size), and rotations (in the range [-5$^\circ$, 5$^\circ$]).

The remaining heads were trained by freezing the backbone and the classification head weights, so only the layers of each of these modules were adjusted. With regard to the pointing and drag heads, the process described in Section \ref{sec:fs} was carried out using the filters learned by the backbone. In the case of the loupe head, a fine-tuning process (during 200 epochs with a batch size of 32) was performed using the dataset composed of images and textual descriptions (see Section \ref{sec:dataset}). 

With regard to the pinch head, sequences labeled as zoom-in, zoom-out or static gestures were used. This module was also trained during 200 epochs with a batch size of 32 but using a special type of data augmentation. In this case, the same transformation was applied to the two frames, which included variations in the speed of the opening and closing gestures (varying the value of $d$ by $\pm$1 frames) and variations in the static gestures (using the same frame or comparing it with the previous and subsequent frame).

The different datasets were divided into two partitions, one for training with 80\% of the samples and the other for test with the remaining samples. Moreover, a small part of the training set (10\%) was used as a validation partition to tune the hyperparameters and to stop training when there was no improvement. These partitions were created by maintaining the same proportion of samples per class. Note also that these same partitions were employed to train and test all the approaches used in this work so as to ensure a fair comparison between them.

% ---------------------------------------------------------------------------
\section{Experiments} 
\label{sec:experiments}

In this section, the different parts of the proposed method are evaluated, starting with the performance of the backbone and the classification head, and continuing with an analysis of the results obtained by each of the specialized heads. In all cases, the results are compared with those of other state-of-the-art methods. It is important to note that all the models were trained and evaluated under the same conditions: they were initialized with the pre-trained weights obtained from ILSVRC dataset, trained during the same number of epochs, with the same datasets described in Section \ref{sec:dataset}, and the same train and test partitions were maintained.

%---------------------------------------------------
\subsection{Evaluation of gesture classification}
\label{sec:classification_results}

In order to assess the performance of the gesture classification methods, three evaluation metrics widely used for this kind of tasks were chosen, Precision, Recall, and $F_1$. Taking one class as positive and the rest as negative, these metrics can respectively be defined as: 

\begin{align}
\text{Precision} &= \frac{TP}{TP + FP}  
\label{eq:precision}
\\
\text{Recall}    &= \frac{TP}{TP + FN} 
\label{eq:recall}
\\
F_1        &= 2 \times \frac{\text{Precision} \cdot \text{Recall}}{\text{Precision} + \text{Recall}}
\label{eq:f1}
\end{align}

\noindent where $TP$, $FP$, and $FN$ denote the number of true positives, false positives, and false negatives, respectively. Since the experiments were conducted as a multi-class problem, we report the results in terms of macro-$F_1$ for a global evaluation, which is calculated as the average of the $F_1$ obtained for each class.

In this part of the method (which detects the initial gesture), we considered three different approaches: 1) categorical CNN architectures with which to classify the gesture in the input image, 2) the use of object detection networks to detect and classify the gesture, and 3) the use of a hand tracking approach based on the MediaPipe library \cite{Hands2020mediapipe}, which detects both the hand and the position of the fingers. 

\textbf{Categorical CNN architectures.} 
With regard to the first type of approximation, 13 representative state-of-the-art CNN topologies (listed in Table \ref{tab:synthetic}) were considered. The reader is referred to the cited works for the implementation details. These topologies were used as backbones, whose final layers were removed and replaced with the classification header described in Section~\ref{sec:classification}. Table~\ref{tab:synthetic} shows the results obtained for the classification task using the synthetically generated dataset (see Section \ref{sec:dataset}). Please recall that all the networks were initialized with the pre-trained weights obtained with the ILSVRC dataset and then trained and evaluated for this synthetic dataset. As will be noted, quite good values are achieved in all cases. The best results are those of Darknet-53 followed by the Xception model, and the worst are those obtained by the VGG architectures.

\begin{table}[ht]
\caption{Comparison of the results obtained with the 13 state-of-the-art CNN topologies considered for the classification of the synthetic dataset. The best results obtained per metric are marked in bold type, while the second-best are underlined.}
\label{tab:synthetic}
\renewcommand{\arraystretch}{.7}
\setlength{\tabcolsep}{10pt}
\centering
\begin{tabular}{lccc}
\toprule
\textbf{Model}      & \textbf{Precision} 	& \textbf{Recall} 		& $\boldsymbol{F_1}$  \\ 
\midrule
SqueezeNet~\cite{iandola2016squeezenet}     & 87.25                 & 86.13                	& 86.69 \\
ResNet-50~\cite{Kaiming2016resnet} 		& 95.74                 & 94.75                	& 95.24 \\
VGG 16~\cite{Simonyan2014}     		& 85.12                 & 67.37                	& 75.21 \\
VGG 19~\cite{Simonyan2014}       	& 84.77                 & 65.40    	& 73.84 \\
Inception v3~\cite{Szegedy2015b}  	& 96.60                 & 96.25                	& 96.42 \\
MobileNet v1~\cite{howard2017mobilenets}    & 97.85                 & 97.83                	& 97.84 \\
MobileNet v2~\cite{sandler2018mobilenetv2}  & 97.24    			& 97.12                	& 97.18 \\
MobileNet v3 \cite{howard2019movilenetv3}	& 94.83    			& 92.99               	& 93.90 \\
EfficientNet-B0 \cite{Tan2019EfficientNet} 	& 97.16    			& 97.05            	& 97.10 \\
EfficientNet-B1 \cite{Tan2019EfficientNet} 	& 97.64    			& 97.41           	& 97.52 \\
DenseNet121 \cite{Huang2017densenet}        & 96.51    			& 96.03              	& 96.27 \\
Xception~\cite{Chollet2016XceptionDL}  	& \underline{98.33}     & \underline{98.30}     & \underline{98.31} \\
Darknet-53~\cite{redmon2018yolov3}   	& \textbf{99.76} 	& \textbf{99.75}        & \textbf{99.75} \\
\bottomrule
\end{tabular}
\end{table}

The weights learned using the synthetic dataset were used to initialize the networks before training with the real image dataset. Table \ref{tab:classsify} shows the results obtained for this second step of the training process. This table also includes a comparison with the result that would be obtained from training without this initialization, that is, initializing with ILSVRC and then training directly with the real dataset. As will be observed, the best results are again obtained with the Darknet-53 architecture followed by Xception. Note that this initialization helps improve the results by an average of more than 3 \%, and if these results are analyzed individually, by up to almost 10 \% in the case of EfficientNet-B0 and 5.5 \% for Darknet-53.

\begin{table}[ht]
\caption{Results obtained for the categorical classification of the real image dataset after initializing the 13 state-of-the-art CNN topologies considered using the weights learned with the synthetic dataset. This table also shows the result obtained when not applying this initialization, that is, starting only with the weights learned from ILSVRC. The best results obtained per metric are marked in bold type, while the second-best are underlined.}
\label{tab:classsify}
\renewcommand{\arraystretch}{.7}
\setlength{\tabcolsep}{10pt}
\centering
\begin{adjustbox}{width = \textwidth, keepaspectratio}
\begin{tabular}{lccclccc}
\toprule
& \multicolumn{3}{c}{\textbf{No initialization}}                    && \multicolumn{3}{c}{\textbf{Synthetic initialization}}      \\  \cmidrule{2-4}\cmidrule{6-8}
\textbf{Model}                              & \textbf{Precision} 	& \textbf{Recall} 		& $\boldsymbol{F_1}$       && \textbf{Precision} 	& \textbf{Recall} 		& $\boldsymbol{F_1}$  \\ 
\midrule
SqueezeNet~\cite{iandola2016squeezenet}     & 69.13             	& 67.08                	& 68.09             && 71.23               & 67.95                 & 69.55 \\
ResNet-50~\cite{Kaiming2016resnet} 		& 80.26             	& 78.98                	& 79.61             && 81.78             	& 80.07                	& 80.92  \\
VGG16~\cite{Simonyan2014}     		& 76.72             	& 73.58                	& 75.12             && 78.45             	& 76.29                	& 77.35 \\
VGG19~\cite{Simonyan2014}       		& 79.19             	& 78.33    	& 78.76             && 81.20             	& 79.90   	& 80.54 \\
Inception v3~\cite{Szegedy2015b}  	& 75.68             	& 73.33                	& 74.49             && 78.62            	& 75.50                	& 77.03 \\
MobileNet v1~\cite{howard2017mobilenets} 	& 82.72             	& 82.75                	& 82.73             && 83.96             	& 83.04                	& 83.50 \\
MobileNet v2~\cite{sandler2018mobilenetv2} 	& 83.44    			& 82.42                	& 82.93             && 84.68	& 86.60	& 84.12 \\
MobileNet v3 \cite{howard2019movilenetv3}   & 84.13    			& 79.86               	& 81.94             && 85.24	& 83.27	& 84.24 \\
EfficientNet-B0 \cite{Tan2019EfficientNet}  & 67.29    		& 61.09               	& 64.04 	&& 78.50	& 69.86	& 73.93 \\
EfficientNet-B1 \cite{Tan2019EfficientNet}  & 75.15    		&  69.50           	& 72.21             && 78.59	& 75.14	& 76.86 \\
DenseNet121 \cite{Huang2017densenet}	& 82.88    			& 79.29               	& 81.05             && 85.90	& 81.43	& 83.61 \\
Xception~\cite{Chollet2016XceptionDL}      	& \underline{87.01} 	& \underline{86.25}     & \underline{86.63} && \underline{91.42} 	& \underline{90.21}     & \underline{90.81}\\
Darknet-53~\cite{redmon2018yolov3}   	& \textbf{89.80} 	& \textbf{87.64}        & \textbf{88.71}    && \textbf{95.31} 	& \textbf{93.14}        & \textbf{94.21} \\
\midrule
\textbf{Average}                            & 79.49	& 76.93                 & 78.18	&& 82.68               & 79.95                 & 81.28 \\
\bottomrule
\end{tabular}
\end{adjustbox}
\end{table}

\textbf{Object detection networks.} 
The second approach evaluated for the classification of the initial gesture was the use of object detection networks, for which four alternatives were compared: Faster R-CNN (FRCNN)~\cite{ren2015:fasterrcnn}, RetinaNet~\cite{Lin2017RetinaNet}, YOLO v3~\cite{redmon2018yolov3}, and SelAE~\cite{Gallego2018selae}. Table \ref{tab:categorical-summary} shows the results obtained when employing these methods, and compares them with the two best results obtained previously by the classification networks (see Table \ref{tab:classsify}). The object detection methods return the position of the gestures in the image (the labeling used for this process is described in Section~\ref{sec:dataset}). Since these networks can return multiple predictions, the bounding box predicted with the highest confidence was selected. In order to evaluate the result obtained, it is necessary to differentiate between whether or not the ground truth contains a gesture. When it does, the prediction is considered TP if its IoU with the ground truth is greater than 0.5, FP when it is less than 0.5, or FN in the case of not returning any prediction. The opposite applies for the gesture ``None'', which is considered to be TP when there is no prediction and FP when there is. 

Table \ref{tab:categorical-summary} shows that, of all the object detection approaches, YOLO v3 obtains the best result, followed by SelAE and RetinaNet. However, if we compare them with the methods specifically trained for categorical classification, Darknet-53 (the backbone used by YOLO v3 itself) still obtains the best score.

\begin{table}[ht]
\caption{Summary of the results obtained by the different approaches considered for the gesture classification task. The best result for each metric is marked in bold type, and the second best result is underlined.}
\label{tab:categorical-summary}
\renewcommand{\arraystretch}{.7}
\setlength{\tabcolsep}{10pt}
\centering
\begin{tabular}{llccc}
\toprule
\textbf{Approach}  		
	& \textbf{Model}                                & \textbf{Precision} 	& \textbf{Recall} 		& $\boldsymbol{F_1}$  \\ 
\midrule
\multirow{2}{*}{Categorical} 
	&	Xception~\cite{Chollet2016XceptionDL}      	& 91.42 	& 90.21                 & 90.81 \\
&	Darknet-53~\cite{redmon2018yolov3}   	& \textbf{95.31} 	& \textbf{93.14}        & \textbf{94.21} \\
\midrule
\multirow{4}{*}{Object detection} 
	&	FRCNN \cite{ren2015:fasterrcnn}  		& 78.71                 & 70.34           		& 74.29       \\
	&	RetinaNet \cite{Lin2017RetinaNet}   	& 86.25                 & 83.29  		& 84.74       \\
	&   SelAE \cite{Gallego2018selae}			& 88.17	& 86.21	& 87.18         \\
	&	YOLO v3 \cite{redmon2018yolov3}       	& \underline{94.30}     & \underline{93.13}     & \underline{93.71} \\ 
\midrule
Hand tracking 
	&	MediaPipe~\cite{Hands2020mediapipe}         & 76.81      		& 77.94      			&  77.37     \\ 
\bottomrule
\end{tabular}
\end{table}

\textbf{Hand tracking approach.} 
This table also includes the result obtained using the third approach: the detection of gestures based on the hand tracking method provided by MediaPipe Hands~\cite{Hands2020mediapipe}. This method detects 21 3D keypoints corresponding to the joints of the fingers of a hand from a single RGB image. We evaluated different approaches to classify gestures using this information, such as kNN (k-Nearest Neighbor) or SVM either directly on the 21 keypoints or by accumulating the joint angles of each finger. The latter (using the sum of angles and SVM) was that which obtained the best results, and was, therefore, the one that was finally included in the comparison. However, as will be observed in the table, the results of this method are not competitive if we compare them with those of the other approximations (with the exception of FRCNN). These worse results are owing to the fact that, in many cases, MediaPipe does not detect the hand correctly. These results also coincide with a recent work \cite{Amaliya2021} in which the performance of this method is compared with other approaches for the recognition of sign language. 

\begin{figure}[ht]
	\centering
	\includegraphics[width=.7\textwidth]{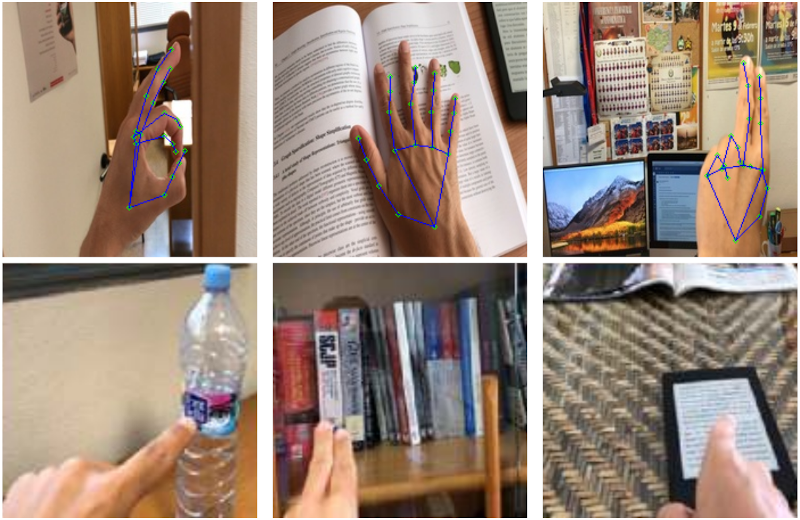}
\caption{Some examples of the detections made by MediaPipe. The first row shows correct detections and the second row shows the cases in which the method failed to detect the hand.} 
\label{fig:mediapipe_results}
\end{figure}

Figure~\ref{fig:mediapipe_results} shows some samples of the detections made by this method. The first row shows correct detections and the second row shows the cases in which it has not been able to detect the hand. As will be noted, these failures occur when the hand is partially occluded (i.e., only the fingers or part of the hand can be seen), which is quite common in the proposed application owing to the position of the camera.

\subsubsection{Multi-objective optimization problem}

Another important aspect to consider is the efficiency of the method selected. Since the more parameters models have, the slower the performance and the greater the storage space required, it is necessary to reach a trade-off between the efficiency of the network and its accuracy. However, these criteria are, quite often, contradictory, since an improvement to one of them usually entails a worsening of the other. From this point of view, this task can be seen as a Multi-objective Optimization Problem (MOP) in which two functions are meant to be optimized simultaneously. 

The most common means employed to deal with problems of this nature is that of resorting to the concept of \textit{non-dominance}: one solution is said to dominate another if, and only if, it is better or equal in each objective function, and at least strictly better in one of them. The best solutions (there may be more than one) are, therefore, those that are non-dominated. In the MOP framework, the strategies within this set define the so-called Pareto frontier and can be considered the best without having to define any order among them~\cite{mie99}.

\begin{figure}[ht]
\centering
\includegraphics[width=.9\textwidth]{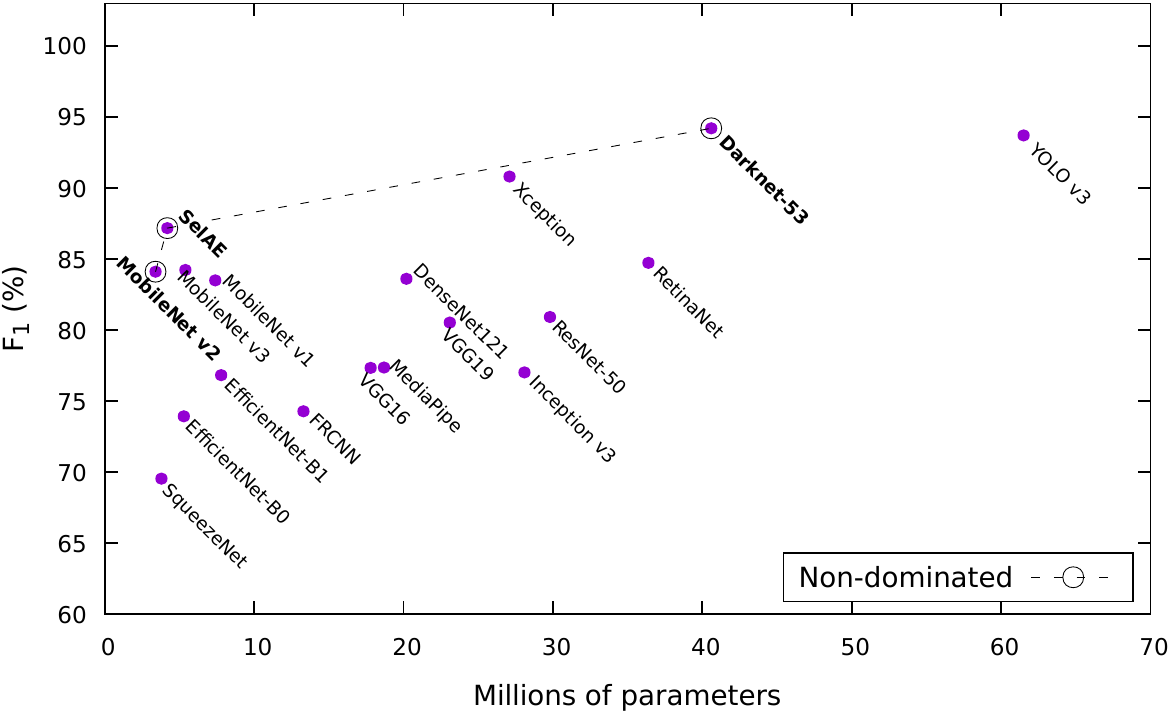}
\caption{Analysis of $F_1$ and efficiency as a Multi-objective Optimization Problem (MOP). Non-dominated elements are highlighted.}
\label{fig:mop-v1}
\end{figure}

Assuming this MOP scenario, Figure~\ref{fig:mop-v1} shows the results obtained by the different methods evaluated in the previous section, in which each point is a 2-dimensional value defined by its $F_1$ and the number of parameters of the corresponding topology. As can be seen, the best results for both criteria (i.e., the \textit{non-dominated} elements) are obtained by MobileNet v2 (with the lowest number of parameters), Darknet-53 (with the best $F_1$), and SelAE (as an intermediate solution). However, and as previously argued, it is essential to obtain high precision in the first step of the proposed method, and it was for this reason that DarkNet-53 was selected for the backbone, since it obtains an $F_1$ that is 7.03 \% higher than the next non-dominated result (SelAE). In addition, this architecture performs well in current mobile devices. According to the tests conducted on a Samsung A51 and a Huawei P30 lite, an average response time of between 3 and 4 FPS was obtained.

%---------------------------------------------------
\subsubsection{Temporal consistency}

\begin{figure}[ht]
	\centering
\includegraphics[width=.65\textwidth]{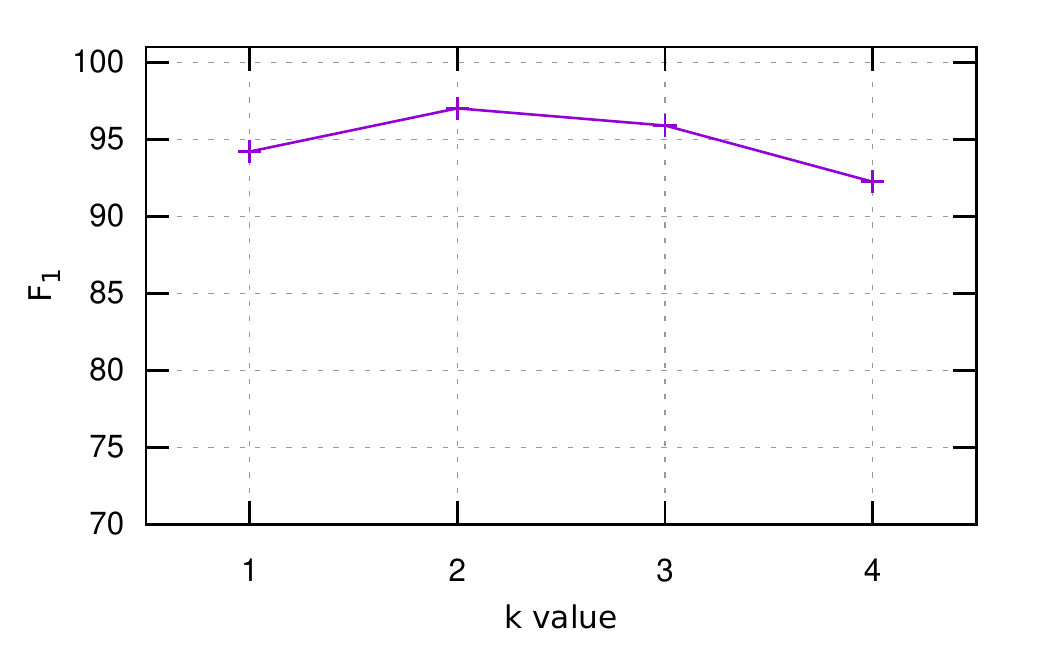}
\caption{Evaluation of the $k$ parameter in the temporal consistency for the classification task using the Darknet-53 backbone.} 
	\label{fig:window_frames}
\end{figure}

Another important part of the proposed architecture that had to be evaluated was the temporal consistency module (see Figure \ref{fig:scheme}) and the effect of the value selected for the $k$ parameter. This parameter makes it possible to control the number of frames with the same response that must elapse for the response to be valid. Figure \ref{fig:window_frames} shows a graph in which this value is studied in the range $[1, 4]$ for the classification task and using the Darknet-53 backbone. As can be seen, the result improves when it is set to 2 or 3 frames, since this prevents frames with isolated errors from occurring. However, if this value is increased, the result starts to worsen, since it has to wait many frames in order to change the response, and it consequently makes mistakes in all the transitions between gestures. The decision was, therefore, made to set the value of $k$ at 2, since at most it generates an erroneous frame between transitions, and in return it improves the $F_1$ from 94.21 \% to 97.03 \%.

%---------------------------------------------------
\subsection{Evaluation of pointing and drag gestures}
\label{sec:pointing_evaluation}

Once the main configuration of the architecture had been established, the results obtained by each of the specialized heads were analyzed. This section focuses on the heads used for the pointing and drag gestures, starting with an analysis of the methodology proposed for these actions (see Section \ref{sec:fs}), which is then compared with other architectures. In all cases, the dataset with fingertip labeling for the point and drag gestures was used (see Section \ref{sec:dataset}).

The results were also evaluated using the $F_1$ metric (Equation \ref{eq:f1}), but in this case we considered the objects (i.e., the fingertips) whose location was correctly detected. This was done by calculating the bounding box of the predicted objects ($P$), which was then matched with the bounding box of the ground truth ($B$) with which it had a higher IoU (using Equation \ref{eq:iou}). A predicted bounding box $P$ was considered to be correctly detected if $\text{IoU}(P, B) \geq \lambda$. We established $\lambda = 0.5$, a threshold value commonly used in this type of tasks, and calculated the metric $F_1$ considering the correct detections to be TP (i.e., when their $\text{IoU}$ was greater than $\lambda$), the wrong detections to be FP (i.e., when a $P$ did not overlap with any $B$ with a $\text{IoU}$ greater than $\lambda$), and those cases in which a ground truth object was not detected were considered to be FN. Note that if multiple detections of the same object were predicted, only the first was counted as positive while the others were counted as negative.

First, an evaluation of the parameters of the proposed methodology (see Section \ref{sec:fs}) will be carried out, starting with the selected layer (variable $l$ in Equation \ref{eq:Fc}). This will be done by setting the remaining parameters to an initial configuration ($|{\cal F}^c|=3$, $\beta=0.6$, and a square structuring element of size $s=5\times{5}$), and by varying only the selected layer $l$. Three representative network architectures from those previously evaluated (Darknet-53, DenseNet121 and MobileNet v3) will be considered in this analysis, although this adjustment was made for all networks, as will be shown later. 

\begin{figure}[ht]
\centering
\begin{subfigure}[t]{0.49\textwidth}
	\includegraphics[width=1\linewidth]{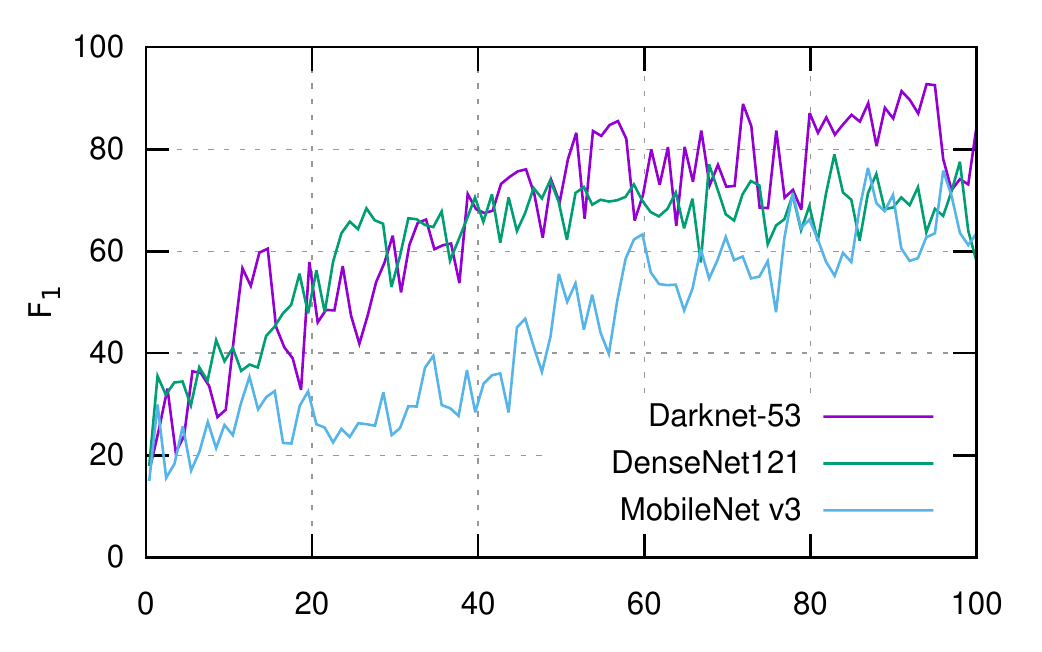}
	\caption{Network depth (\%)}
	\label{fig:fs-layers}
\end{subfigure}
\begin{subfigure}[t]{0.49\textwidth}
	\includegraphics[width=1\linewidth]{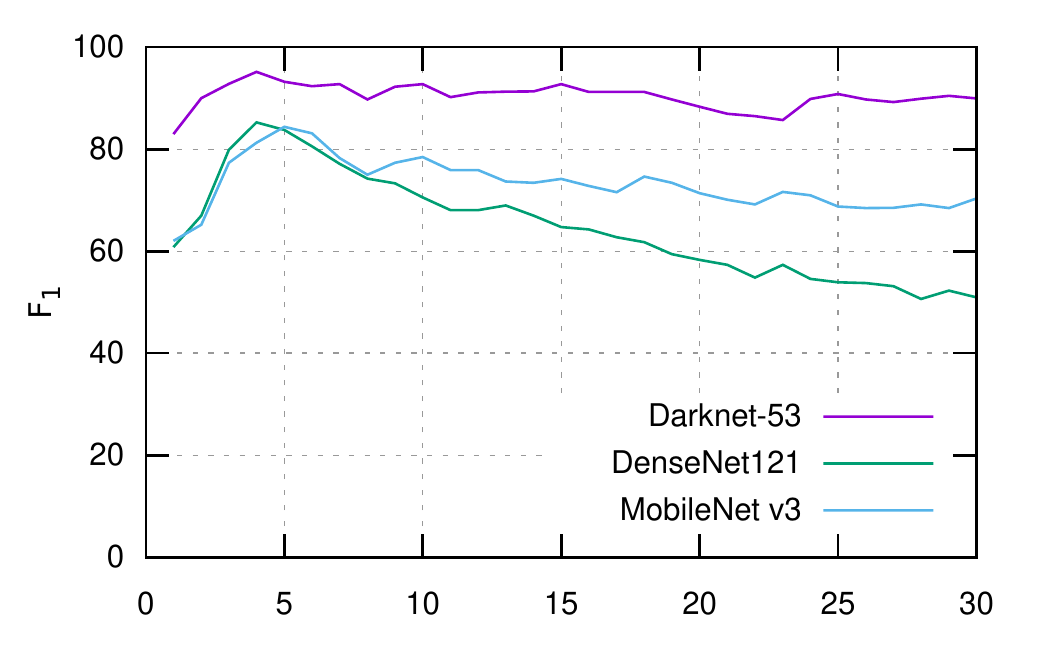}
	\caption{\# filters ($|{\cal F}^c|$)}
	\label{fig:fs-nbfilters}
\end{subfigure}
\begin{subfigure}[t]{0.49\textwidth}
	\includegraphics[width=1\linewidth]{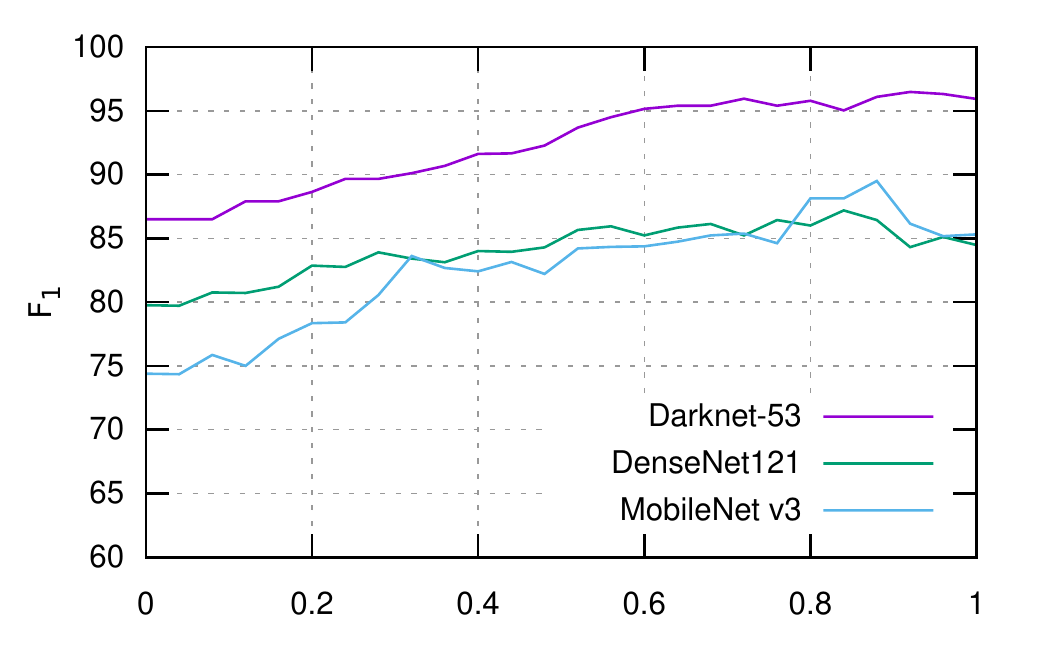}
	\caption{Threshold $\beta$}
	\label{fig:fs-thbeta}
\end{subfigure}
\begin{subfigure}[t]{0.49\textwidth}
	\includegraphics[width=1\linewidth]{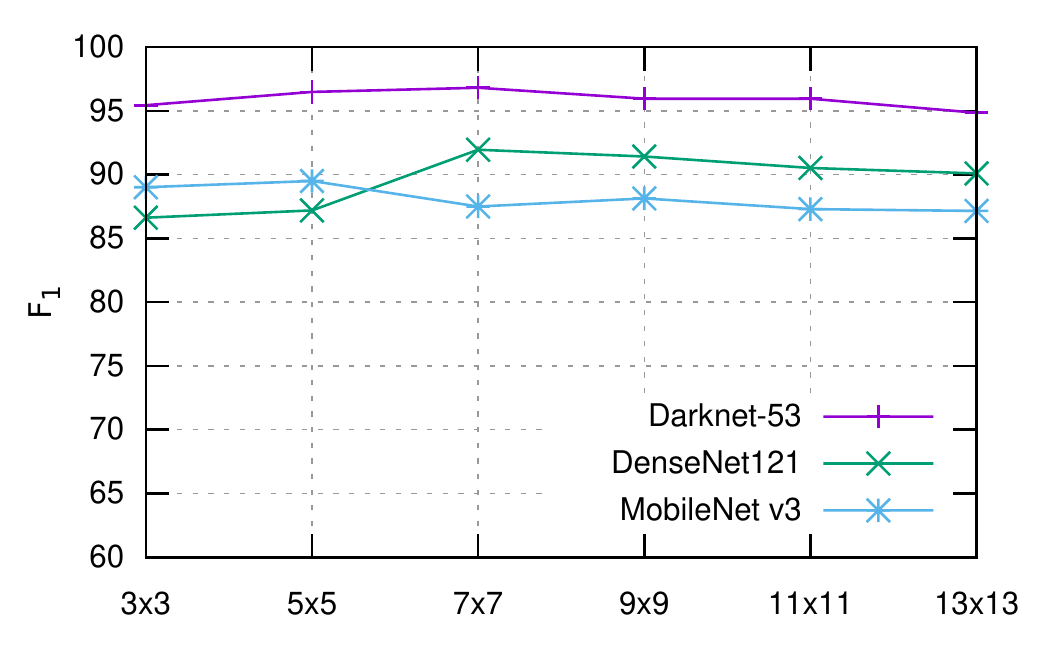}
	\caption{Structuring element $s$ size}
	\label{fig:fs-kernel}
\end{subfigure}
\caption{Localization results ($F_1$ \%) obtained by varying (a) the layer of the network from which the filters are selected (given in percentages with respect to 100\% of the total network depth), (b) the number of filters in the set $|{\cal F}^c|$, (c) the threshold $\beta$, and (d) the structuring element $s$ size.}
\end{figure}

Figure \ref{fig:fs-layers} shows the influence of the CNN layer selected in order to predict the localization. This was done by computing the result obtained with all the layers from the network models considered, while the remaining parameters were set to the aforementioned values. Since each network has a different number of layers, in this figure we represent the result as a function of the layer depth, where 100\% of the depth signifies the last layer of the network. As will be observed, the results in the first part (more or less up to 50\% of depth) were not good. However, as expected, better localization results were obtained in the last part of the networks (from 50\% of depth), from which higher-level features are generally learned. The ``\textit{conv2d\_49}'' layer was, therefore, eventually selected for Darknet-53, ``\textit{conv5\_block6\_2\_conv}'' for DenseNet121, and ``\textit{re\_lu\_35}'' for MobileNet v3 (the full network architectures can be consulted in the corresponding papers).

Another important variable that had to be analyzed was the number of filters selected in order to obtain the localization, that is, the size of the set $|{\cal F}^c|$ in Equation \ref{eq:FS}, which can be adjusted by modifying the threshold value $\alpha$. In this experiment, we used the best layers previously selected: $\beta=0.6$ and $s=5\times5$. Figure \ref{fig:fs-nbfilters} shows the results obtained by varying the size of this set. As will be noted, a maximum is obtained when using between 3 and 6 filters, and the best results are obtained with 4 filters for Darknet-53 and DenseNet121, and with 5 filters for MobileNet v3.

Another parameter that had to be analyzed was the value of the threshold $\beta$ (see Eqs. \ref{eq:P} and \ref{eq:FS}). As before, we set the remaining parameter values to the best ones found and varied only this parameter in the range $[0, 1]$. Figure \ref{fig:fs-thbeta} shows that better results are obtained with higher values for this threshold, i.e., when selecting only those pixels with the highest activations. The specific values selected for each network are: $\beta=0.92$ for Darknet-53, $\beta=0.84$ for DenseNet121, and $\beta=0.88$ for MobileNet v3.

Finally, we also analyzed the influence of the size of the structuring element $s$ (see Eqs. \ref{eq:P} and \ref{eq:FS}) that is used for the dilation of the result obtained from the activation of the filters before calculating the bounding box with the position of the detected objects. The influence of this parameter was assessed by varying the size of the structuring element between $3\times3$ and $13\times13$, and setting the remaining parameters to the best ones found in the previous experiments. Figure \ref{fig:fs-kernel} shows the result of this analysis. As can be seen, the result remains fairly stable when varying this parameter, and improves only slightly with a kernel size of $7\times7$ for Darknet-53 and DenseNet121, and $5\times5$ for MobileNet v3. 

Having analyzed the different parameters of the proposed method and determined the best configuration, the results obtained are now compared with those of other state-of-the-art methods, including the four object detection networks evaluated previously (FRCNN \cite{ren2015:fasterrcnn}, RetinaNet \cite{Lin2017RetinaNet}, YOLO v3 \cite{redmon2018yolov3}, and SelAE \cite{Gallego2018selae}) and the result obtained when applying FS to the other categorical networks considered previously. For this comparison, we show the average value of the IoU obtained, along with the Average Precision (AP), given that these metrics are widely used to evaluate object detection methods, as occurs in the PASCAL VOC challenge. The most recent PASCAL challenge AP metric has been used (by interpolating all the points rather than using a fixed set of uniformly-spaced recall values)~\cite{everingham2015pascal}. This metric calculates the mean value in the recall interval $[0,1]$, which is equivalent to the area under the curve (AUC) of the Precision-Recall curve (PRC). The mAP is calculated by averaging the AP obtained for each class.

Table \ref{tab:fs-comparison} shows the results of this comparison. As can be seen, the method proposed in order to carry out this task (Darknet-53 + FS) is that which obtains the best results, followed by YOLO v3 (the architecture to which the backbone employed belongs). In general, the object detection approaches obtain quite good results. However, in addition to slightly improving the result, FS is a more efficient solution since it does not add any processing layer to the network, it simply takes advantage of the activations of the filters already calculated by the backbone in order to perform this detection.

\begin{table}[ht]
\renewcommand{\arraystretch}{.7}
\setlength{\tabcolsep}{10pt}
\caption{Comparison of the results obtained using the proposed approach (Darknet-53 + FS) and other state-of-the-art methods, including object detection methods, and the result obtained after applying FS to the other categorical networks considered previously. The best results for each metric are marked in bold type, while the second-best are underlined.}
\label{tab:fs-comparison}
\centering
\begin{tabular}{llcc}
\toprule
\textbf{Approach}   & \textbf{Method}   & \textbf{Avg(Iou)}     & \textbf{mAP} \\ 
\midrule
\multirow{4}{*}{Object detection}  
&   FRCNN               & 77.95        & 85.46      \\
&   RetinaNet           & 83.35        & 90.08     \\ 
&   YOLO v3             & \underline{84.71}        & \underline{95.63}      \\
&   SelAE               & 83.38        & 91.21     \\
\midrule
\multirow{13}{*}{Filter selection}  
&   SqueezeNet + FS     & 65.25        & 70.12      \\
&   ResNet-50 + FS      & 74.74        & 83.99     \\
&   VGG16 + FS          & 71.07        & 79.46     \\
&   VGG19 + FS          & 71.91        & 81.77     \\
&   Inception v3 + FS   & 58.89        & 78.30     \\
&   MobileNet v1 + FS   & 71.13        & 84.28     \\
&   MobileNet v2 + FS   & 74.51        & 85.61     \\
&   MobileNet v3 + FS   & 75.40        & 88.93     \\ 
&   EfficientNetB0 + FS & 77.40        & 85.25     \\
&   EfficientNetB1 + FS & 79.57        & 91.25     \\
&   DenseNet121 + FS    & 80.26        & 91.20     \\
&   Xception + FS       & 83.66        & 92.14     \\
&   Darknet-53 + FS     & \textbf{85.77}   & \textbf{96.32}     \\
\bottomrule
\end{tabular}
\end{table}

\begin{figure}[!ht]
\centering
\includegraphics[width=.63\textwidth]{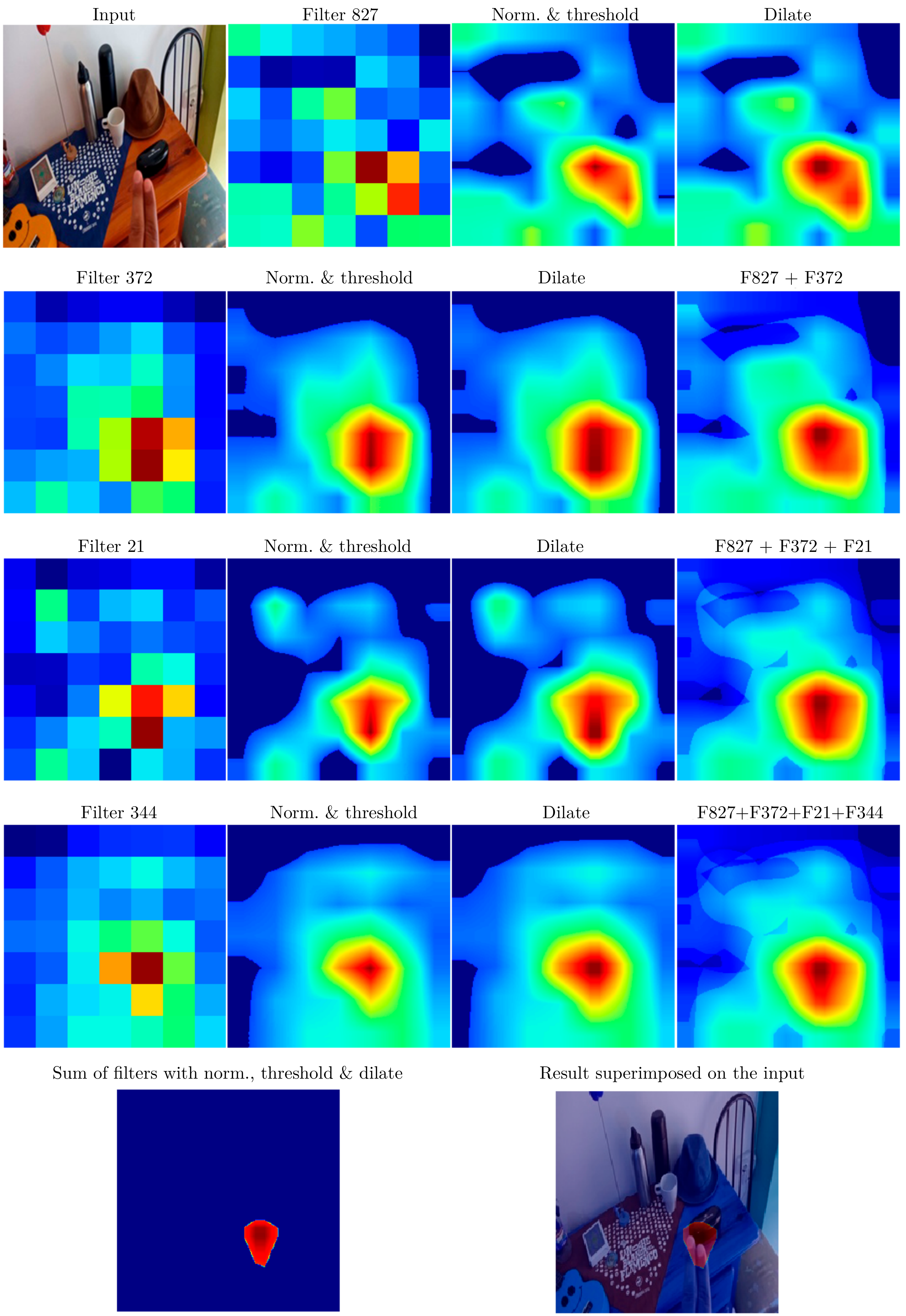}
\caption{Example of the process performed to calculate the location of the fingertips using the proposed approach (Darknet-53 + FS). The first image in the first row is the input. The intermediate results obtained for each of the four filters selected are shown, along with their incremental sum with the previous filters. The last row shows the final result, including an overlay of the predicted localization with the input image. A higher activation value is indicated in dark red.}
\label{fig:fs-example}
\end{figure}

Figure \ref{fig:fs-example} shows an example of the filters obtained for an input image, along with the process of adding up the result until the final prediction is attained. The first row of this image shows the input frame and the process carried out for the first filter, while the second, third and fourth rows show, in addition to the process performed on the filter, the result of the incremental sum with the previous filters.

%---------------------------------------------------
\subsection{Loupe gesture evaluation}

The descriptions generated by the head dedicated to the loupe gesture were evaluated by employing the widely adopted Bi-Lingual Evaluation Understudy (BLEU) metric \cite{papineni2002bleu}. It is generally used to assess the quality of machine-generated sentences by comparing them with reference sentences in problems related to language generation, image captioning, text summarizing, or speech recognition, among others. The output of this metric is in the range $[0, 1]$, where values closer to 1 represent more similar texts. A score of 1 indicates that the sentences are the same. However, it is not necessary to attain this value for the text to be correct. The use of n-grams of a length of between 1 and 4 were considered for the calculation of this metric. This length refers to the number of words in a row that have to match, signifying that the length of 4 (denoted as BLEU-4) would be the most challenging. Also note that for this experiment, in addition to the images with the loupe gesture, the pointing gesture and the images without gesture were also evaluated. These last two cases were added to consider further evaluation samples and also to assess them in case they also have to generate a captioning in the final application. 

Table \ref{tab:bleu-results} shows the results of this evaluation, in which the proposed method (Darknet-53 + captioning) is compared to the \textit{merge-model} method proposed by Marc Tanti et al.~\cite{tanti2018put} (on which the specialized head of our proposal is based) and with the result that would be obtained when exchanging Darknet-53 for DenseNet121 or MobileNet v3. The case of independently training the network formed by Darknet-53 followed by the layers used by the captioning head is also considered. In the latter case, the network is initialized with the weights pre-trained for ILSVRC and then trained (without freezing the backbone) with the dataset created for this task. 

As will be observed in the table, the best result is obtained by the proposed method, but when trained independently. However, this result is only slightly better than that obtained when considering the whole proposed architecture (result of the last row). This small improvement does not, therefore, justify the use of an independent network to carry out this task, since this would suppose a considerable increase in the resources required. The worst result is obtained by the merge-model method, perhaps because it considers a simpler backbone (VGG19). As a reference, the results reported by Marc Tanti et al.~\cite{tanti2018put} for the Flikr8k dataset are 0.191, 0.287, 0.424, 0.611 for BLEU-4, BLEU-3, BLEU-2, and BLEU-1, respectively.

\begin{table}[ht]
\centering
\caption{Comparison of the results obtained (in terms of BLEU) for the image captioning task.}
\label{tab:bleu-results}
\begin{tabular}{lcccc}
\toprule
\textbf{Method}                     & \textbf{BLEU-4} & \textbf{BLEU-3} & \textbf{BLEU-2}   & \textbf{BLEU-1} \\ 
\midrule
Merge-model \cite{tanti2018put}     & 0.1518          & 0.1864          & 0.3151            & 0.4987        \\ 
DenseNet121 + captioning            & 0.1997          & 0.3117          & 0.3904            & 0.5882        \\ 
MobileNet v3 + captioning           & 0.1803          & 0.2921          & 0.3477            & 0.5614        \\ 
Independent backbone + captioning   & 0.2163          & 0.3236          & 0.4408            & 0.6135        \\ 
\midrule
Proposed method (Darknet-53 + captioning)    & \textbf{0.2088} & \textbf{0.3181} & \textbf{0.4390}   & \textbf{0.6027}       \\ 
\bottomrule
\end{tabular}
\end{table}

\begin{figure}[!ht]
	\centering
	\includegraphics[width=.99\textwidth]{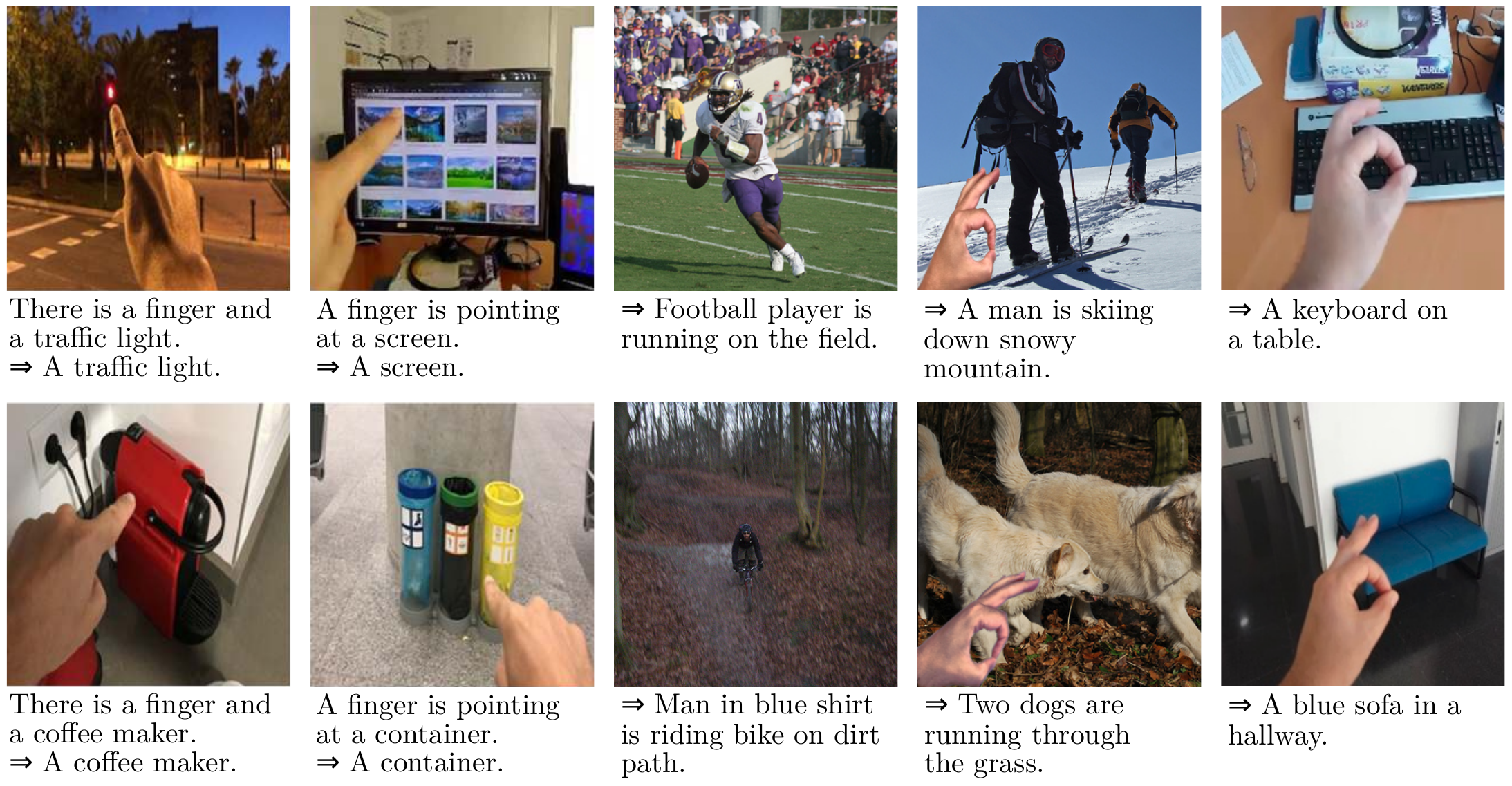}
	\caption{Some evaluation examples of the captioning model using the proposed hand gesture dataset. The first two columns show those cases in which the description mentions the hand or the finger, and the result obtained after the post-processing step.} 
	\label{fig:captioning_results}
\end{figure}

Figure \ref{fig:captioning_results} shows ten examples of the captions generated for our dataset. The first two columns of these examples include the cases in which the texts generated mention the hand or the fingers and the result obtained after post-processing them. As will be noted, the method generates descriptions that correctly detail the scenes and the objects that appear in them.

%---------------------------------------------------
\subsection{Evaluation of pinch gesture}

Finally, the pinch gesture was evaluated. As indicated in the methodology, this is a dynamic gesture that may change in order to indicate whether the user wishes to zoom in, zoom out, or maintain the current zoom level. As in the previous sections, the Precision, Recall and $F_1$ metrics were employed to assess the performance of the detection of these actions. 

For this evaluation, the proposed approach was compared with the use of MediaPipe and with the same network architecture but trained separately (i.e., considering only the backbone and pinch head) and exclusively for this gesture. In this latter case, the backbone was initialized with the weights obtained for ILSVRC, the layers used for the head specialized in the pinch gesture were added and the whole network was fine-tuned for the dataset prepared for this purpose. In the case of MediaPipe, the distance between the thumb and index fingers was calculated in order to determine which gesture was being made.  This made it possible to distinguish among the dynamic gestures performed: zooming in when the distance increases, zooming out when the distance decreases, and maintaining the zoom level when the distance remains stable (allowing a small threshold of $\pm3$ px of variation). 

Table \ref{tab:results-pinch} shows the results of this comparison. As will be observed, MediaPipe obtains the lowest scores since, as previously argued, this method is quite dependent on the visibility of the hand. The proposed approach achieves an $F_1$ of 90.64 \%, which is only 0.97 \% less than the result obtained by this same architecture but trained specifically for this gesture. This experiment shows that the proposed training process does not entail a notable deterioration in the result, but does in return allow the achievement of an efficient system for the simultaneous processing of different actions. This small difference would not, therefore, justify the use of a parallel network to process this gesture, since this would imply a reduction in efficiency and an increase in the resources required.

\begin{table}[ht]
\centering
\caption{Comparison of the results obtained for the pinch gesture.}
\label{tab:results-pinch}
\begin{tabular}{lccc}
\toprule
\textbf{Approach}           & \textbf{Precision}    & \textbf{Recall}   & $\boldsymbol{F_1}$ \\ 
\midrule
\textbf{MediaPipe}          & 63.50                 & 63.57             & 63.53         \\
\textbf{Independent backbone 
+ pinch head}	& 93.27	& 90.01	& 91.61         \\
\midrule
\textbf{Proposed method}    & 92.06                 & 89.27             & 90.64         \\
\bottomrule
\end{tabular}
\end{table}

% ---------------------------------------------------------------------------
\section{Conclusions}
\label{sec:conclusions}

This paper proposes an interactive system for mobile devices controlled by hand gestures, whose objective is to assist people with visual impairments. This system allows users to interact with the device using simple static and dynamic hand gestures, each of which triggers a different action, such as describing the scene or the object pointed to, zooming, etc. The method also optimizes the resources required to perform different tasks, signifying that the system can be embedded in mobile devices. This has been done by employing an efficient multi-head neural network that uses the same features extracted by a common backbone to perform the different actions, which are, moreover, activated only if its corresponding gesture is detected. 

Three different datasets with a total of about 40k images were created to train and evaluate the proposed methodology. The samples were labeled at different levels: category, position of the hands and fingertips, position and category of the objects pointed to, and description of the scenes. The experimentation carried out in each of the steps of the proposed method both attained good results and showed the efficiency of the architecture, resulting in the approximation that obtained the highest precision when adjusting the trade-off between performance and accuracy. The proposed temporal consistency module has proven to improve results by almost 3\% thanks to a simple criterion of continuity in the predictions. When comparing the results of each of the specialized heads with those of other state-of-the-art approaches, including specific options for those same tasks, the best results (or almost the best) are in all cases attained by these specialized heads, thus demonstrating the effectiveness of the proposed architecture, even when compared to specific approaches. Moreover, the architecture has shown a good performance in current mobile devices, with an average response time of between 3 and 4 FPS obtained in tests conducted on a Samsung A51 and a Huawei P30 lite. 

As future work,  the authors intend to carry out a set of user tests with the participation of people with visual impairments in order to evaluate the system and its usability. The objective of this study will be to analyze the proposed interface, the set of gestures considered and the actions performed for each gesture. The proposed architecture allows gestures and actions to be easily added or modified, and the proposed study will, therefore, help adjust the platform to design a more usable interface. The authors also intend to increase the datasets used for the initial classification, and specifically for the detection of objects and the generation of descriptions, in order to expand the variability of the scenes and objects considered.

% ---------------------------------------------------------------------------
%\section*{Acknowledgment}

%\noindent This research was funded ...

%%%%%%%%%%%%%%%%%%%%%%%%%%%%%%%%%%%%%%%%%%%%%%%%%%%%%%%%%%%%%%%%%
% The bibliography
%%%%%%%%%%%%%%%%%%%%%%%%%%%%%%%%%%%%%%%%%%%%%%%%%%%%%%%%%%%%%%%%%

%--------------------------------------------------------------------------------------------
\section*{References}

% \clearpage
% \bibliographystyle{plain}

%% APA style
%\bibliographystyle{model5-names} %\biboptions{authoryear}

%% `Elsevier LaTeX' style
\bibliographystyle{elsarticle-num}

%% IEEE tr.
%\bibliographystyle{ieeetr}

\bibliography{paper}

\end{document}